# Mining meaning from Wikipedia

OLENA MEDELYAN, DAVID MILNE, CATHERINE LEGG and IAN H. WITTEN
University of Waikato, New Zealand

Wikipedia is a goldmine of information; not just for its many readers, but also for the growing community of researchers who recognize it as a resource of exceptional scale and utility. It represents a vast investment of manual effort and judgment: a huge, constantly evolving tapestry of concepts and relations that is being applied to a host of tasks.

This article provides a comprehensive description of this work. It focuses on research that extracts and makes use of the concepts, relations, facts and descriptions found in Wikipedia, and organizes the work into four broad categories: applying Wikipedia to *natural language processing*; using it to facilitate *information retrieval* and *information extraction*; and as a resource for *ontology building*. The article addresses how Wikipedia is being used as is, how it is being improved and adapted, and how it is being combined with other structures to create entirely new resources. We identify the research groups and individuals involved, and how their work has developed in the last few years. We provide a comprehensive list of the open-source software they have produced.

## 1. INTRODUCTION

Wikipedia requires little introduction or explanation. As everyone knows, it was launched in 2001 with the goal of building free encyclopedias in all languages. Today it is easily the largest and most widely-used encyclopedia in existence. Wikipedia has become something of a phenomenon among computer scientists as well as the general public. It represents a vast investment of freely-given manual effort and judgment, and the last few years have seen a multitude of papers that apply it to a host of different problems. This paper provides the first comprehensive summary of this research (up to mid-2008), which we collect under the deliberately vague umbrella of *mining meaning from Wikipedia*. By *meaning*, we encompass everything from concepts, topics, and descriptions to facts, semantic relations, and ways of organizing information. *Mining* involves both gathering meaning into machine-readable structures (such as ontologies), and using it in areas like information retrieval and natural language processing.

Traditional approaches to mining meaning fall into two broad camps. On one side are carefully hand-crafted resources, such as thesauri and ontologies. These resources are generally of high quality, but by necessity are restricted in size and coverage. They rely on the input of experts, who cannot hope to keep abreast of the incalculable tide of new discoveries and topics that arise constantly. Even the most extensive manually created resource—the Cyc ontology, whose hundreds of contributors have toiled for 20 years—has limited size and patchy coverage [Sowa 2004]. The other extreme is to sacrifice quality for quantity and obtain knowledge by performing large-scale analysis of unstructured text. However, human language is rife with inconsistency, and our intuitive understanding of it cannot be entirely replicated in rules or trends, no matter how much data they are based upon. Approaches based on statistical inference might emulate human intelligence for particular purposes and in specific situations, but cracks appear when generalizing or moving into new domains and tasks.

Wikipedia provides a middle ground between these two camps—quality and quantity—by offering a rare mix of scale and structure. With two million articles and thousands of contributors, it dwarfs any other manually created resource by an order of magnitude in the number of concepts covered, has far greater potential for growth, and offers a wealth of further useful structural features. It contains around 18 Gb of text, and its extensive network of links, categories and infoboxes provide a variety of explicitly defined semantics that other corpora lack. One must, however, keep Wikipedia in perspective. It does not always engender the same level of trust or expectations of quality as traditional resources, because its contributors are largely unknown and unqualified. It is also far smaller and less representative of all human language use than the web as a whole. Nevertheless, Wikipedia has received enthusiastic attention as a promising natural language and informational resource of unexpected quality and utility. Here we focus on research that makes use of Wikipedia, and as far as possible leave aside its controversial nature.

This paper is structured as follows. In the next section we describe Wikipedia's creation process and structure, and how it is viewed by computer scientists as anything from a corpus, taxonomy, thesaurus, or hierarchy of knowledge topics to a full-blown ontology. The next three sections describe different research applications. Section 3 explains how it is being drawn upon for *natural language processing*; understanding written text. In Section 4 we describe its applications for *information retrieval*; searching through documents, organizing them and answering questions. Section 5 focuses on *information extraction and ontology building*—mining text for topics, relations and facts—and asks whether this adds up to Tim Berners-Lee's vision of the Semantic Web. Section 6 documents the people and research groups involved, and the resources they have produced, with URLs. The final section gives a brief overall summary.

## 2 WIKIPEDIA: A RESOURCE FOR MINING MEANING

Wikipedia, one of the most visited sites on the web, outstrips all other encyclopedias in size and coverage. Its English language articles alone are ten times the size of the Encyclopedia Britannica, its nearest rival. But material in English constitutes only a quarter of Wikipedia—it has articles in 250 other languages as well. Co-founder Jimmy

Wales is on record as saying that he aspires to distribute a free encyclopedia to every person on the planet, in their own language.

This section provides a general overview of Wikipedia, as background to our discussions in Sections 3–5. We begin with an insight into its unique editing methods, their benefits and challenges (Section 2.1); and then outline its key structural features, including articles, hyperlinks and categories (Section 2.2). Section 2.3 identifies some different roles that Wikipedia as a whole may usefully be regarded as playing—for instance, as well as an encyclopedia it can be viewed as a linguistic corpus. We conclude in Section 2.4 with some practical information on how to work with Wikipedia data.

## 2.1 The encyclopedic wisdom of crowds

From its inception the Wikipedia project offered a unique, entirely open, collaborative editing process, scaffolded by then-new Wiki software for group website building, and it is fascinating to see how it has flourished under this system. It has effectively enabled the entire world to become a panel of experts, authors and reviewers—contributing under their own name, or, if they wish, anonymously.

In its early days the project attracted widespread skepticism. It was thought that its editing system was so anarchic that it would surely fill up with misconceptions, lies, vanity pieces and other worse-than-useless human output. A piece in The Onion satirical newspaper "Wikipedia Celebrates 750 Years Of American Independence: Founding Fathers, Patriots, Mr. T. Honored"[1] nicely captures this point of view. Moreover, it was argued, surely the ability for anyone to make any change, on any page, entirely anonymously, would leave it ludicrously vulnerable to vandalism, particularly to articles that cover sensitive topics. What if the hard work of 2000 people were erased by one eccentric? And indeed "edit wars" did erupt, though it turned out that some of the most vicious raged over such apparently trivial topics as the ancestry of Freddy Mercury and the true spelling of *yoghurt*. Yet this turbulent experience was channeled into developing ever-more sophisticated Wikipedia policies and guidelines,[2] as well as a more subtle code of recommended good manners referred to as Wikiquette.[3] A self-selecting set of administrators emerged, who performed useful functions such as blocking individuals from editing for periods of time—for instance edit warriors, identified by the fact that they "revert" an article more than three times in 24 hours. Interestingly, the development of these rules was guided by the goal of reaching consensus, just as the encyclopedia's content is. Somehow these processes worked sufficiently well to shepherd Wikipedia through its growing pains, and today it is wildly popular and expanding all the time. Section 2.3.1 discusses its accuracy and trustworthiness as an encyclopedia.

[1] *http://www.theonion.com/content/node/50902*
[2] *http://en.wikipedia.org/wiki/Wikipedia:Policies_and_guidelines*

Wikipedia's editing process can be grounded in the knowledge theory proposed by the 19th Century pragmatist Pierce. According to Pierce, beliefs can be understood as knowledge not due to their prior justification, but due to their usefulness, public character and future development. His account of knowledge was based on a unique account of *truth*, which claimed that true beliefs are those that all sincere participants in a "community of inquiry" would converge on, given enough time. Influential 20th Century philosophers [e.g. Quine 1960] scoffed at this notion as being insufficiently objective. Yet Peirce claimed that there is a kind of person whose greatest passion is to render the Universe intelligible and will freely give time to do so, and that over the long run, within a sufficiently broad community, the use of signs is intrinsically self-correcting [Peirce 1877]. Wikipedia can be seen as a fascinating and unanticipated concrete realization of these apparently wildly idealistic claims.

In this context it is interesting to note that Larry Sanger, Wikipedia's co-founder and original editor-in-chief, had his initial training as a philosopher—with a specialization in theory of knowledge. In public accounts of his work he has tried to bypass vexed philosophical discussions of truth by claiming that Wikipedians are not seeking it but rather a neutral point of view.[4] But as the purpose of this is to support every reader being able to build their own opinion, it can be argued that somewhat paradoxically this is the fastest route to genuine consensus. Interestingly, however, he and the other co-founder Jimmy Wales eventually clashed over the role of expert opinion in Wikipedia. In 2007 Sanger diverged to found a new public online encyclopedia Citizendium[5] in an attempt to "do better" than Wikipedia, apparently reasserting validation by external authority—e.g., academics. Interestingly, although it is early days, Citizendium seems to lack Wikipedia's popularity and momentum.

Wikipedia's unique editing methods, and the issues that surround them, have complex implications for mining. First, unlike a traditional corpus, it is constantly growing and changing, so results obtained at any given time can become stale. Some research strives to measure the degree of difference between Wikipedia versions over time (though this is only useful insofar as Wikipedia's rate of change is itself constant), and assess the impact on common research tasks [e.g. Ponzetto and Strube 2007a]. Second, how are projects that incorporate Wikipedia data to be evaluated? If Wikipedia editors are the only people in the world who have been enthusiastic enough to write up certain topics (for instance, details of TV program plots), how is one to determine 'ground truth' for evaluating applications that utilize this information? The third factor is more of an opportunity than a challenge. The awe-inspiring abundance of manual labor given freely to Wikipedia raises the possibility of a new kind of research project, which would consist in encouraging Wikipedians themselves to perform certain tasks on the researchers' behalf—tasks of a scale the researchers themselves could not hope to achieve. As we will see in Section 5, some have begun to glimpse this possibility, while others continue to view Wikipedia in more traditional "product" rather than "process" terms. At any rate, this research area straddles a fascinating interface between software and social engineering.

---

3 *http://en.wikipedia.org/wiki/Wikipedia:WQT*
4 *http://en.wikipedia.org/wiki/Wikipedia:Neutral_point_of_view*
5 *http://en.citizendium.org*

### 2.2. Wikipedia's structure

Traditional paper encyclopedias consist of articles arranged alphabetically, with internal cross-references to other articles, external references to the academic literature, and some kind of general index of topics. These structural features have been adapted by Wikipedia for the online environment, and some new features added that arise from the Wiki editing process. The statistics presented below were obtained from a version of English Wikipedia released in July 2008.

*2.2.1. Articles:* The basic unit of information in Wikipedia is the *article*. Internationally, Wikipedia contains 10M articles in 250 different languages.[6] The English version contains 2.4M articles (not counting redirects and disambiguation pages, which are discussed below). About 1.8M of these are bona fide articles with more than 30 words of descriptive text and at least one incoming link from elsewhere in Wikipedia. Articles are written in a form of free text that follows a comprehensive set of editorial and structural guidelines in order to promote consistency and cohesion. These are laid down in the Manual of Style,[7] and include the following:

1. Each article describes a single concept, and there is a single article for each concept.
2. Article titles are succinct phrases that resemble terms in a conventional thesaurus.
3. Equivalent terms are linked to an article using redirects (Section 2.2.3).
4. Disambiguation pages present various possible meanings from which users can select the intended article. (Section 2.2.2).
5. Articles begin with a brief overview of the topic, and the first sentence defines the concept and its type.

[6] *http://en.wikipedia.org/wiki/Wikipedia*
[7] *http://en.wikipedia.org/wiki/Wikipedia:Manual_of_Style*

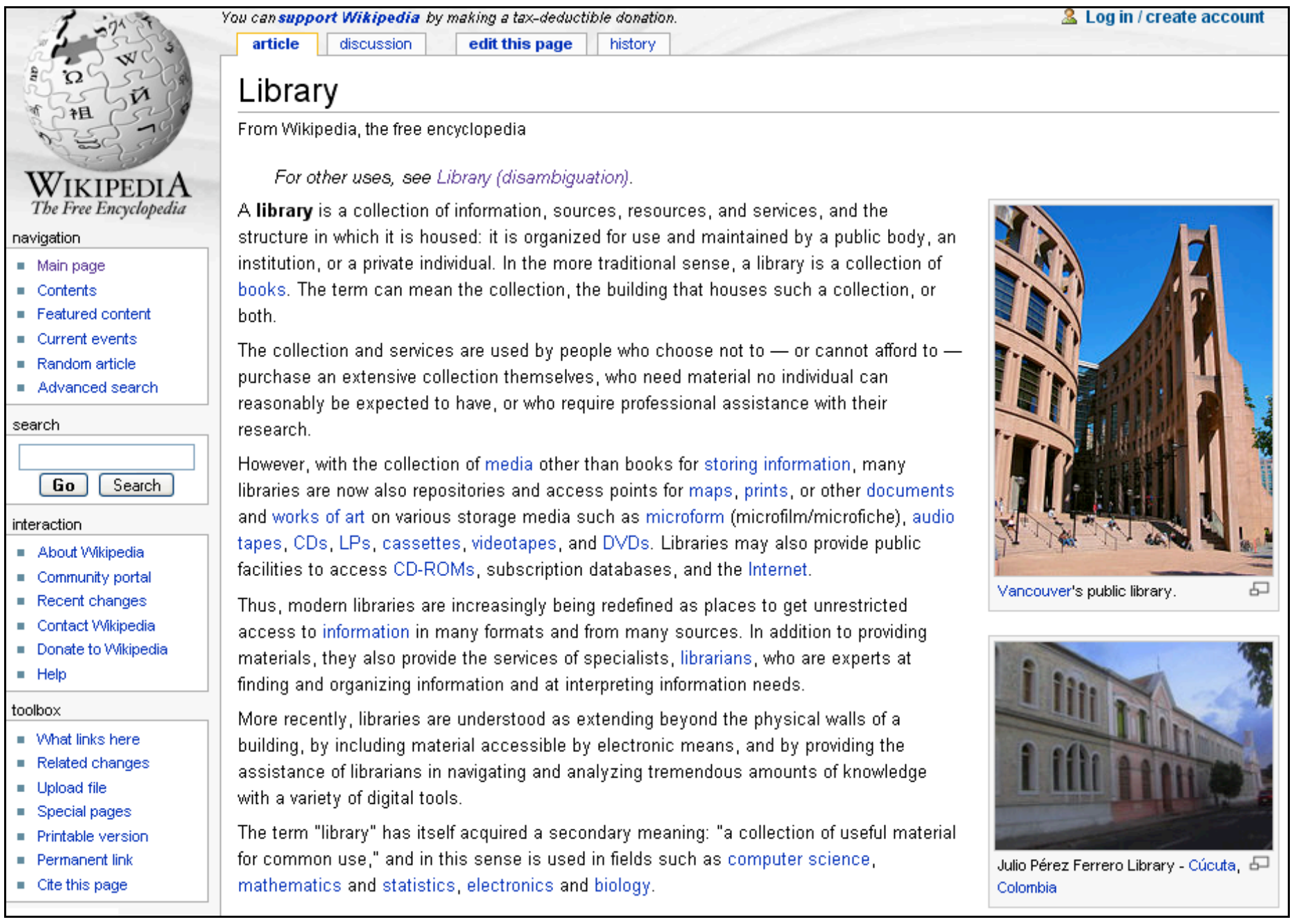

Figure 1. Wikipedia article on *Library*.

6. Articles contain hyperlinks that express relationships to other articles (Section 2.2.4).

Figure 1 shows a typical article, entitled *Library*. The first sentence describes the concept:

> A **library** is a collection of information, sources, resources, and services: it is organized for use and maintained by a public body, an institution, or a private individual.

Here the article's title is the single word *Library*, but titles are often qualified by appending parenthetical expressions. For example, there are other articles entitled *Library* (*computing*), *Library* (*electronics*), and *Library* (*biology*). Wikipedia distinguishes capitalization when it is relevant: the article *Optic nerve* (the nerve) is distinguished from *Optic Nerve* (the comic book).

*2.2.2. Disambiguation pages:* Instead of taking readers to an article named by the term, as *Library* does, the Wikipedia search engine sometimes takes them directly to a disambiguation page where they can click on the meaning they want. These pages are identified by invoking certain templates (discussed in Section 2.2.6) or assigning them to certain categories (Section 2.2.5), and often contain (*disambiguation*) in their title.

The English Wikipedia contains 100,000 disambiguation pages. The first line of the *Library* article in Figure 1 ("For other uses …") links to a disambiguation page that lists *Library* (*computing*), *Library* (*electronics*), *Library* (*biology*), and further senses. Brief scope notes accompany each sense to assist users. For instance, *Library* (*computer science*) is “a collection of subprograms used to develop software.” The articles themselves serve as detailed scope notes. Disambiguation pages are helpful sources of information concerning homonyms.

*2.2.3. Redirects:* A redirect page is one with no text other than a directive in the form of a *redirect* link. There are about a dozen for *Library* and just under 3M in the entire English Wikipedia. They encode pluralism (*libraries*), technical terms (*bibliotheca*), common misspellings (*libary*), and other variants (*reading room, book stack*). As noted above, the aim is to have a single article for each concept and define redirects to link equivalent terms to the article’s preferred title. As we will see, this helps with mining because it is unnecessary to resolve synonymy using an external thesaurus.

*2.2.4. Hyperlinks:* Articles are peppered with hyperlinks to other articles. The English Wikipedia contains 60M in total, an average of 25 per article. They provide explanations of topics discussed, and support an environment where serendipitous encounters with information are commonplace. Anyone who has browsed Wikipedia has likely experienced the feeling of being happily lost, browsing from one interesting topic to the next, encountering information that they would never have searched for.

Because the terms used as anchors are often couched in different words, Wikipedia’s hyperlinks are also useful as an additional source of synonyms not captured by redirects. *Library*, for example, is referenced by 20 different anchors including *library*, *libraries*, and *biblioteca*. Hyperlinks also complement disambiguation pages by encoding polysemy; *library* links to different articles depending on the context in which it is found. They also give a sense of how well-known each sense is; 84% of *library* links go to the article shown in Figure 1, while only 13% go to *Library* (*computing*). Furthermore, since internal hyperlinks indicate that one article relates to another in some respect, this fundamental structure can be mined for meaning in many interesting ways—for example, to capture the associative relations included in standard thesauri (Section 5.1).

*2.2.5. Category structure:* Authors are encouraged to assign categories to their articles. For example, the article *Library* falls in the category *Book Promotion*. Authors are also encouraged to assign the categories themselves to other more general categories; *Book Promotion* belongs to *Books*, which in turn belongs to *Written Communication*. These categorizations, like the articles themselves, can be modified by anyone. There are almost 400,000 categories in the English Wikipedia, with an average of 19 articles and two subcategories each.

Categories are not themselves articles. They are merely nodes for organizing the articles they contain, with a minimum of explanatory text. Often (in about a third of cases), categories correspond to a concept that requires further description. In these cases they are paired with an article of the same name: the category *Libraries* is paired with the article *Library*, and *Billionaires* with *Billionaire*. Other categories, such as *Libraries by country*, have no corresponding articles and serve only to organize the content. For clarity, in this paper we indicate categories in the form *Category:Books* unless it is obvious that we are not talking about an article.

The goal of the category structure is to represent information hierarchy. It is not a simple tree-structured taxonomy, but a graph in which multiple organization schemes coexist. Thus both articles and categories can belong to more than one category. The category *Libraries* belongs to four: *Buildings and structures*, *Civil services*, *Culture* and *Library and information science*. The overall structure approximates an acyclic directed graph; all relations are directional, and cycles are rare—although they sometimes occur. (According to Wikipedia's own guidelines, cycles are generally discouraged but may sometimes be acceptable. For example, *Education* is a field within *Social Sciences*, which is an *Academic discipline*, which belongs under *Education.* In other words, you can educate people about how to educate).

The category structure is a relatively recent addition to the encyclopedia, and less visible than articles. They likely receive much less scrutiny than articles, and have been criticized as haphazard, incomplete, inconsistent, and rife with redundancy [Chernov et al. 2006; Muchnik et al. 2007]. Links represent a wide variety of types and strengths of relationships. Although there has been much cleanup and the greatest proportion of links now represent class membership (*isa*), there are still many that represent other semantic relations (e.g. parthood), and merely thematic associations between entities—as well as meta-categories used for editorial purposes, such as *Disambiguation*. Thus *Category:Pork* currently contains, among others, the categories *Domestic Pig, Bacon Bits, Religious Restrictions on the Consumption of Pork*, and *Full Breakfast*. We will see in Section 5 that these issues have not prevented researchers from innovatively mining the category structure for a range of purposes.

*2.2.6. Templates and infoboxes:* Templates are reusable structures invoked to add information to other pages in an efficient fashion. Wikipedia contains 174,000 different templates, which have been invoked 23M times. They are commonly used to identify articles that require attention; e.g. if they are biased, poorly written, or lacking citations. They can also define pages of different types, such as disambiguation pages or featured (high quality) articles. A common application is to provide navigational links, such as the *for other uses* link shown in Figure 1.

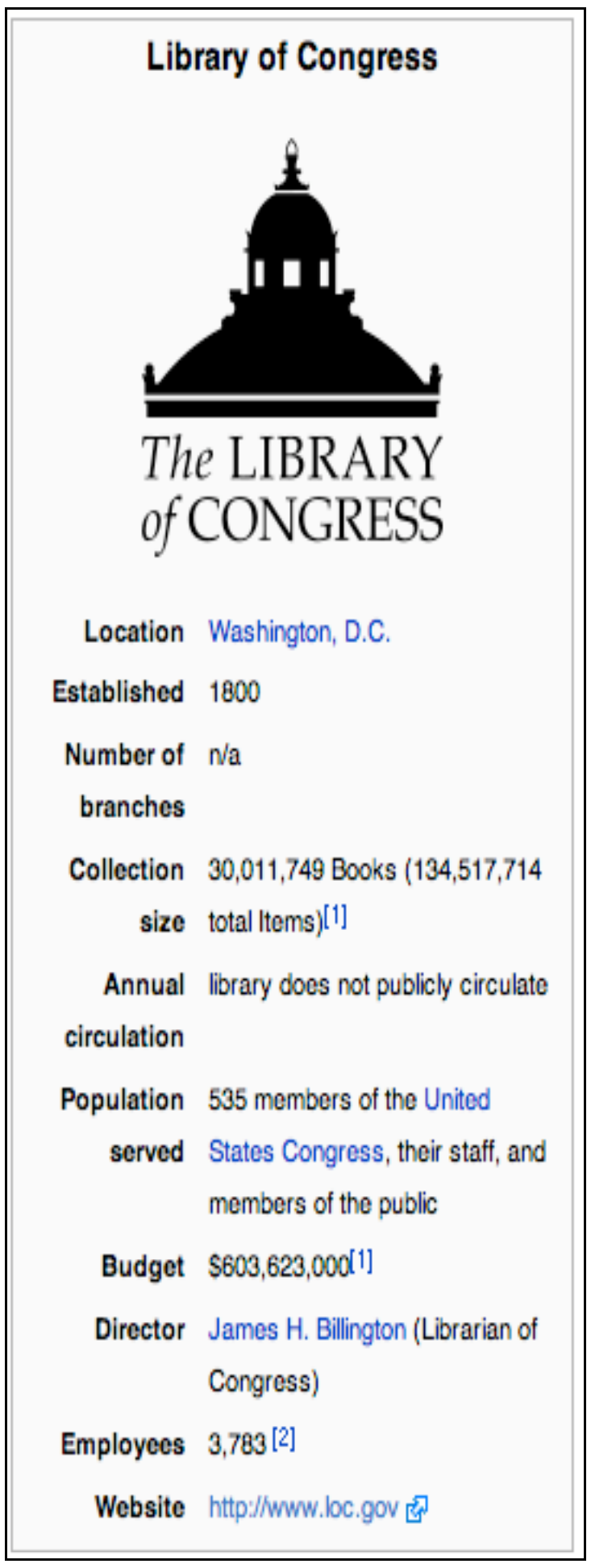

Figure 2. Infobox for the Library of Congress

An infobox is a special type of template that displays factual information in a structured uniform format. Figure 2 shows one from the article on the *Library of Congress*. It was created by invoking the *Infobox Library* template and populating its fields, such as *location* and *collection size.* There are about 3,000 different infobox templates that are used for anything from animal species to strategies for starting a game of chess, and the number is growing rapidly.

The infobox structure could be improved in several simple ways. Standard representations for units would allow quantities to be extracted reliably. Different attribute names are often used for the same kind of content. Data types could be associated with attribute values, and language and unit tags used when information can be expressed in different ways (e.g. Euro and USD). Many Wikipedia articles use tables for structured information that would be better represented as templates [Auer and Lehmann 2007]. Despite these problems, much meaningful and machine-interpretable information can be extracted from Wikipedia templates. This is discussed further in Section 5.3.

*2.2.7. Discussion pages:* A *discussion* tab at the top of each article takes readers to its *Talk* page, representing a forum for discussions (often longer than the article itself) as to how it might be criticized, improved or extended. For example, the talk page of the *Library* article*, Talk:Library*, contains the following observations, among many others:

> **location?**
> Libraries can also be found in churches, prisons, hotels etc. Should there be any mention of this? – *Daniel C. Boyer 20:38, 10 Nov 2003*
> Libraries can be found in many places, and articles should be written and linked. A wiki article on libraries can never be more of a summary, and will always be expandable – *DGG 04:18, 11 September 2006*

There are talk pages for other aspects of Wikipedia's structure, such as templates and categories, as well as user talk pages that editors use to communicate with each other. These pages are a unique and interesting feature of Wikipedia not replicated in traditional encyclopedias. They have been mined to determine quality metrics of Wikipedia edits [Emigh et al. 2005; Viégas et al. 2007], but have not been yet employed for any other tasks discussed in this survey—perhaps because of their unstructured nature.

*2.2.8. Edit histories:* To the right of the *discussion* tab is a *history* tab that takes readers to an article's editing history. This contains the name or pseudonym of every editor, with the changes they made. We can see that the *Library* article was created on 9 November 2001 in the form of a short note—which, in fact, bears little relationship to the current version—and has been edited about 1500 times since. Recent edits add new links and new entries to lists; indicate possible vandalism and its reversal; correct spelling mistakes; and so on.

Analyzing editing history is an interesting research area in its own right. For example, Viégas [2004] mines history pages to discover collaboration patterns. Nelken and Yamangil [2008] discuss ways of utilizing them as a corpus for extracting lexical errors called *eggcorns*, e.g. <*rectify*, *ratify*>, as well as phrases that can be dropped to compress sentences—a useful component of automatic text summarization. It is natural to ask whether the content of individual articles converges in some semantic sense, staying stable despite continuing edits. Thomas and Amit [2007] call the information in a Wikipedia article "justified" if, after going through the community process of discussion, repeated editing, and so on, it has reached a stable state. They found that articles do, in general, become stable, but that it is difficult to predict where a given article is in its journey towards maturity. They also point out that although information in an article's edit history might indicate its likely quality, mining systems invariably ignore it.

Table 1 breaks down the number of different pages and connections in the English version at the time of writing. There are almost 5.5M pages in the section dedicated to articles. Most are redirects. Many others are disambiguation pages, lists (which group related articles but do not provide explanatory text themselves) and stubs (incomplete articles with fewer than 30 words or at least one incoming link from elsewhere in

| **Articles and related pages** | 5,460,000 |
|---|---|
| redirects | 2,970,000 |
| disambiguation pages | 110,000 |
| Lists and stubs | 620,000 |
| bona-fide articles | 1,760,000 |

| **Categories** | 390,000 |
|---|---|

| **Templates** | 174,000 |
|---|---|
| infoboxes | 3,000 |
| other | 171,000 |

| **Links** | |
|---|---|
| between articles | 62,000,000 |
| between category and subcategory | 740,000 |
| between category and article | 7,270,000 |

Table 1. Content of English Wikipedia.

Wikipedia). Removing all these leaves about 1.8M bona-fide articles, each with an edit history and most with some content on their discussion page. The articles are organized into 400,000 different categories and augmented with 170,000 different templates. They are densely interlinked, with 62M connections—an average of 25 incoming and 25 outgoing links per article.

### 2.3. Perspectives on Wikipedia

Wikipedia is a rich resource with several different broad functionalities, and subsequent sections will show how researchers have developed sophisticated mining techniques with which they can identify, isolate and utilize these different perspectives.

*2.3.1 Wikipedia as an encyclopedia:* The first and most obvious usage for Wikipedia is exactly what it was intended as: an encyclopedia. Ironically, this is the very application that has generated most doubt and cynicism. As noted above, the open editing policy has led many to doubt its authority. Denning et al. [2005] review early concerns, and conclude that, while Wikipedia is an interesting example of large-scale collaboration, its use as an information source is risky. Their core argument is the lack of formal expert review procedures, which gives rise to two key issues: accuracy within articles, and bias of coverage across them.

Accuracy within articles is investigated by Giles [2005], who compares randomly selected scientific Wikipedia articles with their equivalent entries in Encyclopedia Britannica. Both sources were equally prone to significant errors, such as misinterpretation of important concepts. More subtle errors, however, such as omissions or misleading statements, were more common in Wikipedia. In the 41 articles reviewed there were 162 mistakes in Wikipedia versus 123 for Britannica. Britannica Inc. attacked Giles' study as "fatally flawed"[8] and demanded a retraction; Nature defended itself and declined to retract.[9] Ironically, while Britannica's part in the debate has been polemical and plainly biased, Wikipedia provides objective coverage on the controversy in its article on *Encyclopedia Britannica*.

Several authors have developed metrics that evaluate the quality of Wikipedia articles based on such features as: number of authors, number of edits, internal and external linking, and article size, e.g. Lih [2004] and Wilkinson and Huberman [2007]; article stability, e.g. Dondio et al. [2006]; and the amount of conflict an article generates, e.g. Kittur [2007]. Emigh and Herring [2005] perform a genre analysis on Wikipedia using corpus linguistic methods to determine "features of formality and informality," and claim that its post-production editorial control produces entries as standardized as those in traditional print encyclopedias. Viégas et al. [2007] claim that overall coordination and organization, one of the fastest growing areas of Wikipedia, ensures great resilience to malicious editing despite high traffic; they highlight in particular the role played by discussion pages.

[8] *http://www.corporate.britannica.com/britannica_nature_response.pdf*
[9] *http://www.nature.com/press_releases/Britannica_response.pdf*

So much for accuracy. A second issue is *bias of coverage*. Wikipedia is edited by volunteers, who naturally apply more effort to describing topics that pique their interest. For example, there are 600 different articles dedicated to *The Simpsons* cartoon. In contrast, there are half as many pages about the namesake of the cartoon's hero, the Greek poet *Homer*, and all the literary works he created and inspired. Some researchers have analyzed Wikipedia's coverage systematically. Halavais and Lackaff [2008], for example, investigated whether the particular enthusiasms of volunteer editors produce excessive coverage of certain topics by comparing topic-distribution in Wikipedia with that in *Books In Print*, and with a range of printed scholarly encyclopedias. They measure this using a Library of Congress categorization of 3000 randomly-chosen articles and find Wikipedia's coverage remarkably representative, except for law and medicine. Milne et al. [2006] compare Wikipedia with an agricultural thesaurus and identify a bias towards concepts that are general or introductory, and therefore more relevant to "everyman." Lih [2004] shows that Wikipedia's content, and therefore bias, is driven to a large extent by the press.

*2.3.2. Wikipedia as a corpus:* Large text collections are useful for creating language models that capture particular characteristics of language use. For example, the language in which a text is written can be determined by analyzing the statistical distribution of the letter n-grams it contains [Cavnar and Trenkle 1994], whereas word co-occurrence statistics are helpful in tasks like spelling correction [Mays et al. 1991]. Aligned corpora in different languages are extremely useful for machine translation [Brown et al. 1993], where extensive coverage and high quality of the corpus are crucial factors for success. While the web has enabled rapid acquisition of large text corpora, their quality leaves much to be desired due to spamming and the varying format of websites. In particular, manually annotated corpora and aligned multilingual corpora are rare and in high demand.

Wikipedia provides a plethora of well-written and well-formulated articles—several gigabytes in the English version alone—that can easily be separated from other parts of the website. The Simple Wikipedia is a reduced version that contains easier words and shorter sentences, intended for people learning English. The absence of complex sentences makes automatic linguistic processing easier, and some researchers focus on this version for their experiments [Ruiz-Casado et al. 2005a; Toral and Muñoz 2006].

Many researchers take advantage of Wikipedia's host of definitions for question answering (Section 4.3) and automatic extraction of semantic relations (Section 5.1). Section 2.2.8 mentions how Wikipedia history pages can be used as a corpus for training text summarization algorithms, as well as for determining the quality of the articles themselves.

Wikipedia also contains annotations in the form of targeted hyperlinks. Consider the following two sentences from the article about the Formula One team named McLaren.

1. The [[Kiwi (people)|Kiwi]] made the team's Grand Prix debut at the 1966 Monaco race.
2. Original McLaren [[Kiwi|kiwi]] logo; a New Zealand icon.

In the first case the word *kiwi* links to *Kiwi* (*people*); in the second to *Kiwi*, the article describing the bird. This mark-up is nothing more or less than word sense annotation. Mihalcea [2007] shows that Wikipedia is a full fledged alternative to manually sense-tagged corpora. Section 3.2 discusses research that makes use of these annotations for word sense disambiguation and computing the semantic similarity between words.

Although the exploration of Wikipedia as a source of multilingual aligned corpora has only just begun, its links between description of concepts in different languages have been exploited for cross-language question answering [Ferrández et al. 2007] and automatic generation of bilingual dictionaries [Erdmann et al. 2008]. This is further discussed in Section 3.4, while Section 4.2 investigates Wikipedia's potential for multilingual information retrieval.

*2.3.3 Wikipedia as a thesaurus:* There are many similarities between the structure of traditional thesauri and the ways in which Wikipedia organizes its content. As noted, each article describes a single concept, and its title is a succinct, well-formed phrase that resembles a term in a conventional thesaurus. If article names correspond to manually defined terms, links between them correspond to relations between terms, the building blocks of thesauri. The international standard for thesauri (ISO 2788) specifies four kinds of relation:

- Equivalence: USE, with inverse form USE FOR
- Hierarchical: broader term (BT), with inverse form narrower term (NT)
- Any other kind of semantic relation (RT, for related term).

Wikipedia redirects provide precisely the information expressed in the equivalence relation. They are a powerful way of dealing with word variations such as abbreviations, equivalent expressions and synonyms. Hierarchical relations (broader and narrower terms) are reflected in Wikipedia's category structure. Hyperlinks between articles capture other kinds of semantic relation. (Restricting consideration to mutual cross-links eliminates many of the more tenuous associations.)

As we will see, researchers compare Wikipedia with manually created domain-specific thesauri and augment them with knowledge from it (Section 3.2.3). Redirects turn out to be very accurate and can safely be added to existing thesauri without further checking. Wikipedia also has the potential to contribute new topics and concepts, and can be used as a source of suggestions for thesaurus maintenance. Manual creation of scope notes is a labor-intensive aspect of traditional thesauri. Instead, the first paragraph of a Wikipedia article can be extracted as a description of the topic, backed up by the full article should more explanation be required. Finally, Wikipedia's multilingual nature allows thesauri to be translated into other languages.

*2.3.4. Wikipedia as a database:* Wikipedia contains a massive amount of highly structured information. Several projects (notably DBpedia, discussed in Section 5.3) extract this and store it in formats accessible to database applications. The aim is two-fold: to allow users to pose database-style queries against datasets derived from Wikipedia, and to facilitate linkage with other datasets on the web. Some projects even aim to extract database-style facts directly from the text of Wikipedia articles, rather than from infoboxes. Furthermore, disambiguation and redirect pages can be turned into a relational database that contains tables for *terms*, *concepts*, *term concept relationships* and *concept relationships* [Gregorowicz and Kramer 2006].

Another idea is to bootstrap fact extraction from articles by using the content of infoboxes as training data and applying machine learning techniques to extract even more infobox-style information from the text of other articles. This allows infoboxes to be generated for articles that do not yet have them [Wu and Weld 2007]. Related techniques can be used to clean up the underlying infobox data structure, with its proliferation of individual templates.

*2.3.5 Wikipedia as an ontology:* Articles can be viewed as ontology elements, for which the URIs of Wikipedia entries serve as surprisingly reliable identifiers [Hepp et al. 2006]. Of course, true ontologies also require concept nodes to be connected by informative relations, and in Section 5 we will see researchers mine such relations in a host of innovative ways from Wikipedia's structure—including redirects, hyperlinks (both incoming and outgoing, as well as the anchor text), category links, category names and infoboxes, and even raw text, as well as experimenting with adding relations to and from other resources such as WordNet and Cyc. From this viewpoint Wikipedia is arguably by far the largest ontological structure available today, with its Wiki technology effectively serving as a large-scale collaborative ontology development environment. Some researchers are beginning to mix traditional mining techniques with possibly more far-sighted attempts to encourage Wikipedia editors themselves to develop the resource in directions that might bear ontological fruit.

*2.3.6 Wikipedia as a network structure:* Wikipedia can be viewed as a hyperlinked structure of web pages, a microcosm of the web. Standard methods of analyzing network structure can then be applied [Bellomi and Bonato 2005]. The two most prominent techniques for web analysis are Google's PageRank [Brin and Page 1998] and the HITS algorithm [Kleinberg 1998]. Bellomi and Bonato [2005] applied both to Wikipedia and discerned some interesting underlying cultural biases (as of April 2005). These authors conclude that PageRank and HITS seem to identify different kinds of information. They report that according to the HITS authority metric, space (in the form of political geography) and time (in the form of both time spans and landmark events) are the primary organizing categories for Wikipedia articles. Within these, information tends to be organized around famous people, common words, animals, ethnic groups, political and social institutions, and abstract concepts such as music, philosophy, and religion.

In contrast, the most important articles according to PageRank include a large cluster of concepts tightly related to religion. For example, *Pope*, *God* and *Priest* were the highest-ranking nouns, as compared to *Television*, *Scientific classification*, and *Animal* for HITS. They found that PageRank seemed to transcend recent political events to give a wider historical and cultural perspective in weighting geographic entities. It also tends to bring out a global rather than a Western perspective, both for countries and cities and for historical events. HITS reveals a strong bias towards recent political leaders, whereas people with high PageRank scores tend to be ones with an impact on religion, philosophy and society. It would be interesting to see how these trends have evolved since the publication of this work.

An alternative to PageRank and HITS is the Green method [Duffy 2001], which Ollivier and Senellart [2007] applied to Wikipedia's hyperlink network structure in order to find related articles. This method, which is based on Markov Chain theory, is related to the topic-sensitive version of PageRank introduced by Haveliwala [2003]. Given a target article, one way of finding related articles is to look at nodes with high PageRank in its immediate neighborhood. For this a topic-sensitive measure like Green's is more appropriate than the global PageRank.

The Wikipedia category graph also forms a network structure. Zesch and Gurevych [2007] showed that it is a scale-free, small-world graph, like other semantic networks such as WordNet. They adapted WordNet-based measures of semantic relatedness to the Wikipedia category graph, and found that they work well—at least for nouns. They suggest that this, coupled with Wikipedia's multilingual nature, may enable natural language processing algorithms to be transferred to languages that lack well-developed semantic WordNets.

### 2.4. Obtaining Wikipedia data

Wikipedia is based on the MediaWiki software. As an open source project, its entire content is easily obtainable in the form of large XML files and database dumps that are released sporadically every several days or weeks.[10] The full content (without revision history or images) of the English version occupies 18 Gb of uncompressed data at the time of writing. Section 6 discusses tools for extracting information from these files.

Instead of obtaining the database directly, specialized web crawlers have been developed. Bellomi and Bonato [2005] scanned the *All pages* index section, which contains a full list of the pages exposed on the website. Pages that do not contain a regular article were identified by testing for specific patterns in the URL, and discarded. Wikipedia's administrators prefer the use of the database dumps, however, to minimize the strain on their services.

## 3 SOLVING NATURAL LANGUAGE PROCESSING TASKS

Natural language processing applications fall into two major groups: i) those using symbolic methods, where the system utilizes a manually encoded repository of human language, and ii) statistical methods, which infer properties of language by processing large text corpora. The problem with the former is a dearth of high-quality knowledge bases. Even the lexical database WordNet, which, as the largest of its kind, receives substantial attention [Fellbaum 1998], has been criticized for low coverage—particularly of proper names—and high sense proliferation [Mihalcea and Moldovan 2001; Ponzetto and Strube 2007a]. Initial enthusiasm with statistical methods somewhat faded once they hit an upper performance bound that is hard to beat unless they are combined with symbolic elements [Klavans and Resnik 1996]. Several research groups simultaneously discovered Wikipedia as an alternative to WordNet. Direct comparisons of their performance on the same tasks have shown that Wikipedia can be employed in a similar way and often significantly outperforms WordNet [Strube and Ponzetto 2006]. This section describes research in the four areas to which Wikipedia has been successfully applied: semantic relatedness (Section 3.1), word sense disambiguation (Section 3.2), co-reference resolution (Section 3.3) and multilingual alignment (Section 3.4).

[10] *http://download.wikimedia.org/wikipedia*

## 3.1 Semantic relatedness

Semantic relatedness quantifies the similarity between two concepts, e.g. *doctor* and *hospital*. Budanitsky and Hirst [2001] differentiate between semantic *similarity*, where only predefined taxonomic relations are used to compute similarity, and semantic *relatedness*, where other relations like *has-part* and *is-made-of* are used as well. Semantic relatedness can be also quantified by statistical methods without requiring a manually encoded taxonomy, for example by analyzing term co-occurrence in a large corpus [Resnik 1999; Jiang and Conrath 1997].

To evaluate automatic methods for estimating semantic relatedness, the correlation coefficient between machine-assigned scores and those assigned by human judges is computed. Three standard datasets are available for evaluation:

- Miller and Charles' [1991] list of 30 noun pairs, which we denote by M&C;
- Rubenstein and Goodenough's [1965] 65 synonymous word pairs, R&G,
- [Finkelstein et al. 2002]'s collection of 353 word pairs (WordSimilarity-353), WS-353.

The best pre-Wikipedia result for the first set was a correlation of 0.86, achieved by Jiang and Conrath [1997] using a combination of statistical measures and taxonomic analysis derived from WordNet. For the third, Finkelstein et al. [2002] achieved 0.56 correlation using Latent Semantic Analysis. The discovery of Wikipedia began a new era of competition.

Strube and Ponzetto [2006] and Ponzetto and Strube [2007a] re-calculated several measures developed for WordNet using Wikipedia's category structure. The best performing metric on most datasets was Leacock and Chodorow's [1998] normalized path measure:

$$lch(c_1,c_2) = -\log \frac{length(c_1,c_2)}{2D},$$

where *length* is the number of nodes on the shortest path between nodes $c_1$ and $c_2$ and $D$ is the maximum depth of the taxonomy. WordNet-based measures outperform Wikipedia-based ones on the small datasets M&C and R&G, but on WS-353 Wikipedia wins by a large margin. Combining similarity evidence from Wikipedia and WordNet using a SVM to learn relatedness from the training data yielded the highest correlation score of 0.62 on a designated "testing" subset of WS-353.

Strube and Ponzetto remark that WordNet's sense proliferation was responsible for its poor performance on WS-353. For example, when computing the relatedness of *jaguar* and *stock*, the latter is interpreted in the sense of animals kept for use or profit rather than in the sense of *market*, which people find more intuitive. WordNet's fine sense granularity has been also criticized in word sense disambiguation (Section 3.2.1). The overall conclusion is that Wikipedia can serve AI applications in the same way as hand-crafted knowledge resources. Zesch et al. [2007] perform similar experiments with the German Wikipedia, which they compare to GermaNet on three datasets including the translated M&C. The performance of Wikipedia-based measures was inconsistent, and, like Strube and Ponzetto [2006], they obtained best results by combining evidence from GermaNet and Wikipedia.

Ponzetto and Strube [2007a] investigate whether performance on Wikipedia-based relatedness measures changes as Wikipedia grows. After comparing February 2006, September 2006 and May 2007 versions they conclude that the relatedness measure is robust. There was no improvement, probably because new articles were unrelated to the words in the evaluation datasets. A Java API is available for those wishing to experiment with these techniques [Ponzetto and Strube [2007c].[11]

Gabrilovich and Markovitch [2007] developed Explicit Semantic Analysis (ESA) as an alternative to the well-known Latent Semantic Analysis. They use a centroid-based classifier to map input text to a vector of weighted Wikipedia articles. For example, for *Bank of Amazon* the vector contains *Amazon River*, *Amazon Basin*, *Amazon Rainforest*, *Amazon.com*, *Rainforest*, *Atlantic Ocean Brazil*, etc. To obtain the semantic relatedness between two terms, they compute the cosine similarity of their vectors. This significantly outperforms Latent Semantic Analysis on WS-353, with an average correlation of 0.75 (with the same technique, the Open Directory Project[12] achieves 0.65 correlation, indicating that Wikipedia's quality is greater). This is easily the best result of all of the techniques described in this section. It also has the unique advantage of being equally applicable to individual words, phrases or even entire documents. The mapping developed in this work has been successfully utilized for text categorization (Section 4.5).

While Gabrilovich and Markovitch [2007] use the full text of Wikipedia articles to establish relatedness between terms, Milne [2007] analyses just the internal hyperlinks. To compute the relatedness between two terms, they are first mapped to corresponding Wikipedia articles, then vectors are created containing the links to other Wikipedia articles that occur in these articles. For example, a sentence like *Bank of America is the largest commercial <bank> in the <United States> by both <deposits> and <market capitalization>* contributes four links to the vector. Each link is weighted by the inverse number of times it is linked from other Wikipedia articles—the less common the link, the higher its weight. For example, *market capitalization* receives higher weight than *United States* and thus contributes more to the semantic relatedness.

---

[11] *http://www.eml-r.org/english/research/nlp/download/jwordnetsimilarity.php*
[12] *http://www.dmoz.org*

Disambiguation is a serious challenge for this technique. Strube and Ponzetto [2006] choose the most likely meaning from the order in which entries occur in Wikipedia's disambiguation pages; Gabrilovich and Markovitch [2007] avoid disambiguation entirely by simultaneously associating a term with several Wikipedia articles. However, Milne's [2007] approach hinges upon correct mapping of terms to Wikipedia articles. When terms are manually disambiguated, a correlation of 0.72 is achieved for WS-353. Automatic disambiguation that simply selects whatever meaning produces the greatest similarity score achieves only 0.45, showing that unlikely senses often produce greater similarity than common ones.

Milne and Witten's [2008a] Wikipedia Link-based Measure—an incremental improvement over [Milne, 2007]—disambiguates term mappings automatically using three features. One is the conditional probability of the sense given the term, according to the Wikipedia corpus (discussed further in Section 3.2.1). For example, the term *leopard* more often links to the animal than the eponymous Mac operating system. They also analyze how commonly two terms appear in Wikipedia as a collocation. Finally, they augment the vector-based similarity metric described above by a measure inspired by Cilibrasi and Vitanyi's [2007] Normalized Google Distance, which is based on term occurrences in web pages, but using the links made to each Wikipedia article rather than Google's search results. The semantic similarity of two terms is determined by the sum of these three values—conditional probability, collocation and similarity. This technique achieves 0.69 correlation with human judgments on WS-353, not far off Gabrilovich and Markovitch's [2007] figure for ESA. However, it is far less computationally intensive because only links are analyzed, not the entire text. Further analysis shows that performance is even higher on terms that are well defined in Wikipedia.

| **Method** | **M&C** | **R&G** | **WS-353** |
|---|---|---|---|
| WordNet<br>[Strube and Ponzetto, 2006] | 0.82 | 0.86 | full: 0.36<br>test: 0.38 |
| WikiRelate!<br>[Ponzetto and Strube, 2007] | 0.49 | 0.55 | full: 0.49<br>test: 0.62 |
| ESA<br>[Gabrilovich and Markovitch, 2007] | 0.73 | 0.82 | 0.75 |
| WLVM<br>[Milne, 2007] | n/a | n/a | man: 0.72<br>auto: 0.45 |
| Wikipedia Link-based Measure<br>[Milne and Witten, 2008a] | 0.70 | 0.64 | 0.69 |

Table 2. Overview of semantic relatedness methods.

To summarize, estimating semantic similarity with Wikipedia has been addressed from three distinct angles:

- applying WordNet-based techniques to Wikipedia: Ponzetto and Strube [2006, 2007a] and Zesch et al. [2007];
- using vector model techniques to compare similarity of Wikipedia articles, similar to Latent Semantic Analysis: Gabrilovich and Markovitch [2007];
- exploring Wikipedia's unique feature of hyperlinked article descriptions: Milne [2007], Milne and Witten [2008a].

The approaches are easily compared because they have been evaluated consistently using the same data sets. Table 2 summarizes the results for each similarity metric described in this section. ESA is best, with WLM not far behind and WikiRelate the lowest. The astonishingly high correlation with human performance that these techniques obtain was well out of reach in pre-Wikipedia days. Also, Wikipedia provides relatedness measures for a far larger vocabulary than resources like WordNet.

Semantic similarity measures are useful for a host of tasks in information retrieval, natural language processing, artificial intelligence and other areas. So far the algorithms described here are underutilized, given the large advances in accuracy and vocabulary that they offer. Later sections (e.g. 3.2, 3.3 and 4.5) provide a few examples of their use, but we expect many more applications in the future.

## 3.2 Word sense disambiguation

Techniques for word sense disambiguation—i.e., resolving polysemy—use a dictionary or thesaurus that defines the inventory of possible senses [Ide and Véronis 1998]. Wikipedia provides an alternative resource. Each article describes a concept that is a possible sense for words and phrases that denote it, whether by redirection via a disambiguation page, or

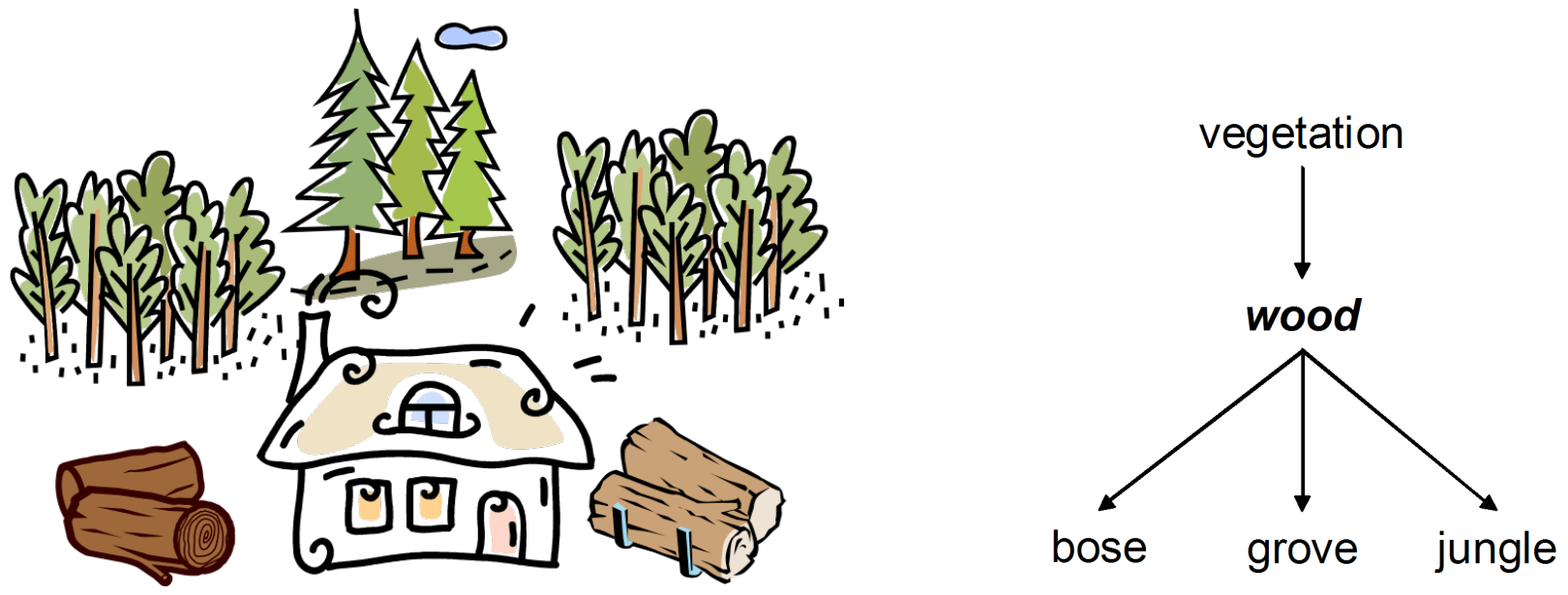

*He could see wood around the house.*

Figure 3. What is the meaning of *wood*?

as anchor text that links to the article.

The terms to be disambiguated may either appear in plain text or in an existing knowledge base (thesaurus or ontology). The former is more challenging because the context is less clearly defined. Consider the example in Figure 3. Even human readers cannot be sure of the intended meaning of *wood* from the sentence alone, but a diagram showing semantically related words in WordNet puts it into context and makes it clear that the meaning is *the trees and other plants in a large densely wooded area* rather than *the hard fibrous lignified substance under the bark of trees*. This highlights the main idea behind disambiguation: identify the context and analyze which of the possible senses fits it best. We first cover techniques for disambiguating phrases in text to Wikipedia articles, then examine the important special case of named entities, and finally show how disambiguation is used to map manually created knowledge structures to Wikipedia.

*3.2.1. Disambiguating phrases in running text.* Discovering the intended senses of words and phrases is an essential stage in every natural language application, otherwise full “understanding” cannot be claimed. WordNet is a popular resource for word sense disambiguation, but success has been mixed [Voorhees 1998]. One reason is that the task is demanding because “linguistic [disambiguation] techniques must be essentially perfect to help” [Voorhees 1998]; another is that WordNet defines senses with such fine granularity that even human annotators struggle to differentiate them [Edmonds and Kilgariff 1998]. The two are related, because fine granularity makes disambiguation more difficult. In contrast, Wikipedia defines only those senses on which its contributors reach consensus, and includes an extensive description of each one rather than WordNet’s brief gloss. Substantial advances have been made since it was discovered as a resource for disambiguation.

Mihalcea [2007] uses Wikipedia articles as a source of sense-tagged text to form a training corpus for supervised disambiguation. They follow the evaluation methodology developed by SIGLEX, the Association for Computational Linguistics’ Special Interest Group on the Lexicon.[13] For each example they collect its occurrences as link anchors in Wikipedia. For example, the term *bar* is linked to *bar* (*establishment*) and *bar* (*music*), each of which corresponds to a WordNet synset—that is, a set of synonymous terms representing a particular meaning of *bar*. The results show that a machine learning approach trained on Wikipedia sentences in which both meanings of *bar* occur clearly outperforms two simple baselines.

This work uses Wikipedia solely as a resource to disambiguate words or phrases into WordNet synsets. Mihalcea and Csomai [2007] go further, using Wikipedia’s content as a sense inventory in its own right. They disambiguate terms—words or phrases—that appear in plain text to Wikipedia articles, concentrating exclusively on "important" concepts. They call this process *wikification* because it simulates how Wikipedia authors manually insert hyperlinks when writing articles. There are two stages: extraction and disambiguation. In the first, terms that are judged important enough to be highlighted as links are identified in the text. Only terms occurring at least five times in Wikipedia are considered, and *likelihood* of a term being a hyperlink is estimated by expressing the number of articles in which a given word or phrase appears as anchor text as a proportion of the total number of articles in which it appears. All terms whose likelihood exceeds a predefined threshold are chosen, which yields an F-measure of 55% on a subset of manually annotated Wikipedia articles.

In the second stage these terms are disambiguated to Wikipedia articles that capture the intended sense. For example, in the sentence *Jenga is a popular beer in the bars of Thailand* the term *bar* corresponds to the *bar* (*establishment*) article. Given a term, those articles for which it is used as anchor text in the Wikipedia are candidate senses. The number of times a term links to a particular sense quantifies the commonness of this sense, a statistic that Mihalcea and Csomai use as a baseline. Their best performing disambiguation algorithm uses a machine learning approach in which Wikipedia's already-annotated articles serve as training data. Features—like part-of-speech tag, local context of three words to the left and right, and their part-of-speech tags—are computed for each ambiguous term that appears as anchor text of a hyperlink. A Naïve Bayes classifier is then applied to disambiguate unseen terms. Csomai and Mihalcea [2007] report an F-measure of 87.7% on 6,500 examples, and go on to demonstrate that linking educational material to Wikipedia articles in this manner improves the quality of knowledge that people acquire when reading the material, and decreases the time taken.

In a parallel development, Wang et al. [2007a] use a fixed-length window to identify terms in a document that match the titles of Wikipedia articles, eliminating matches subsumed by longer ones. They disambiguate the matches using two methods. One works on a document basis, seeking those articles that are most similar to the original document according to the standard cosine metric between TF×IDF-weighted word vectors. The second works on a sentence basis, computing the shortest distance between the candidate articles for a given ambiguous term and articles corresponding to any non-ambiguous terms that appear in the same sentence. The distance metric is 1 if the two articles link to each other; otherwise it is the number of nodes along the shortest path between two Wikipedia categories to which they belong, normalized by the maximum depth of the category taxonomy. This technique is inspired by WordNet based approaches to comparing concepts, and is quite similar to Ponzetto and Strube's [2007a] semantic similarity measure (Section 3.1). The result of disambiguation is the average of the two techniques (if no unambiguous articles are available, the similarity technique is applied by itself). Wang et al. do not compare this method to other disambiguation techniques directly. They do, however, report the performance of text categorization before and after synonyms and hyponyms of matching Wikipedia articles, and their related terms, were added to the documents. The findings were mixed, and somewhat negative.

[13] *http://www.senseval.org*

Medelyan et al. [2008] use Mihalcea and Csomai's [2007] wikification strategy with a different disambiguation technique. Document terms with just one match are unambiguous, and their corresponding articles are collected and used as "context articles" to disambiguate the remaining terms. This is done by determining the average semantic similarity of each candidate article to all context articles identified for the document. The semantic similarity of a pair of articles is obtained from their incoming links as described by Milne and Witten [2008a] (Section 3.1). Account is also taken of the prior probability of a sense given the term, according to the Wikipedia corpus (proposed by Mihalcea and Csomai [2007] as a baseline). For example, the term *jaguar* links to the article *Jaguar cars* in 466 out of 927 cases, thus its prior probability is 0.5. The resulting mapping is the one with the largest product of semantic similarity and prior probability. This achieves an F-measure of 93% on 17,500 mappings in manually annotated Wikipedia articles.

Milne and Witten [2008b] extend both Mihalcea and Csomai [2007] and Medelyan et al. [2008] by applying machine learning to the problems of extracting and disambiguating terms. Disambiguation is improved by using the conditional probability and semantic similarity measures identified in the latter work as features. The use of a decision tree learner and a third feature (which measures the quantity and homogeneity of available context) produces a scheme that can adjust the relative importance of similarity and prior probability from document to document. This raises the F-measure from 93% to 97% on the same data. Topic detection is improved by combining Michalcea and Csomai's link likelihood measure with many other features, including relatedness, generality, disambiguation confidence, and frequency of occurrence. This raises the F-measure from an estimated 48% to 74%.

It is interesting to compare the strategies in the above approaches. Each needs some sort of context to disambiguate a given term to a Wikipedia concept. Some researchers (e.g., Medelyan et al. [2008]) use concepts appearing in the same sentence or text, while others (e.g., Wang et al. [2007a]) build a vector containing all terms of this document. Next, semantic similarity is computed between each candidate meaning and the context using techniques like those described in Section 3.1: traversing the category tree, vector or link analysis. Mihalcea and Csomai [2007] and Milne and Witten [2008b] combine different signals, like similarity, commonness, context and part-of-speech tags, in the final disambiguation classifier trained on Wikipedia's own articles.

Disambiguating terms in running text to articles in Wikipedia can be viewed from another perspective: How can a given text be mapped to a set of relevant Wikipedia concepts?—In other words, which Wikipedia articles are most similar to it? Approaches like ESA (Sections 3.1, 4.5) adopt this perspective, but have not yet been evaluated in the same way as the techniques summarized here. Future research is needed to fill this gap.

So far, the techniques described here have seen minimal application, apart from some explorations into educational support [Mihalcea and Csomai 2007], and topic indexing [Medelyan et al. 2008]. Milne and Witten [2008b] argue that they have enormous potential, since these algorithms cross-reference documents with what is arguably the largest knowledge base in existence, and can provide structured knowledge about any unstructured document. Thus any task that is currently addressed using the bag of words model, or with knowledge obtained from less comprehensive knowledge bases, could likely benefit from these techniques.

*3.2.2. Disambiguating named entities.* Phrases referring to named entities, which are proper nouns such as geographical and personal names, and titles of books, songs and movies, contribute the bulk of our day-to-day vocabulary. Wikipedia is recognized as the largest available collection of such entities. It has become a platform for discussing news, and contributors put issues into encyclopedic context by relating them to historical events, geographic locations and significant personages, thereby increasing the coverage of named entities. Here we describe three approaches that focus specifically on linking named entities appearing in text or search queries to corresponding Wikipedia articles. Section 5.3 summarizes techniques for recognizing named entities in Wikipedia itself.

Bunescu and Paşca [2006] disambiguate named entities in search queries in order to group search results by the corresponding senses. They first create a dictionary of 500,000 entities that appear in Wikipedia, and add redirects and disambiguated names to each one. If a query contains a term that corresponds to two or more entries, they choose the one whose Wikipedia article has the greatest cosine similarity with the query. If the similarity scores are too low they use the category to which the article belongs instead of the article itself. If even this falls below a predefined threshold they assume that no mapping is available. The reported accuracies are between 55% and 85% for members of Wikipedia's *People by occupation* category, depending on the model and experimental data employed.

Cucerzan [2007] identifies and disambiguates named entities in text. Like Bunescu and Paşca [2006], he first extracts a vocabulary from Wikipedia. It is divided into two parts, the first containing surface forms and the second the associated entities along with contextual information. The surface forms are titles of articles, redirects, and disambiguation pages, and anchor text used in links. This yields 1.4M entities, with an average of 2.4 surface forms each. Further <named entity, tag> pairs are extracted from Wikipedia list pages—e.g., *Texas* (*band*) receives a tag *LIST_band name etymologies*, because it appears in the list with this title—yielding a further 540,000 entries. Categories assigned to Wikipedia articles describing named entities serve as tags too, yielding 2.6M entries. Finally a context for each named entity is collected—e.g., parenthetical expressions in its title, phrases that appear as link anchors in the article's first paragraph of the article, etc.—yielding 38M <named entity, context> pairs.

To identify named entities in text, capitalization rules indicate which phrases are surface forms of named entities. Co-occurrence statistics generated from the web by a search engine help to identify boundaries between them (e.g. *Whitney Museum of American Art* is a single entity, whereas *Whitney Museum in New York* contains two). Lexical analysis is used to collate identical entities (e.g., *Mr. Brown* and *Brown*), and entities are tagged with their type (e.g., *location*, *person*) based on statistics collected from manually annotated data. Disambiguation is performed by comparing the similarity of the document in which the surface form appears with Wikipedia articles that represent all named entities that have been identified in it, and their context terms, and choosing the best match. Cucerzan [2007] achieves 88% accuracy on 5,000 entities appearing in Wikipedia articles, and 91% on 750 entities appearing in news stories.

Kazama and Torisawa [2007] recognize and classify entities but do not disambiguate them. Their work resembles the methods described above. Given a sentence, their goal is to extract all n-grams representing Wikipedia articles that correspond to a named entity, and assign a type to it. For example, in the sentence *Rare Jimmy Hendrix song draft sells for almost $17,000* they identify *Jimmy Hendrix* as an entity of type *musician*. To determine the type they extract the first noun phrase following the verb *to be* from the Wikipedia article's first sentence, excluding phrases like *kind of*, *type of*—e.g., *guitarist* in *Jimmy Hendrix was a guitarist*. Recognition is a supervised tagging process based on standard features such as surface form and part of speech tag, augmented with category labels extracted from Wikipedia and a gazetteer. An F-measure of 88% was achieved on a standard set of 1000 training and 220 development and testing documents.

To summarize this research on disambiguating named entities, Cucerzan [2007] and Kazama and Torisawa [2007] report similar performance, while Bunescu and Paşca's [2006] results seem slightly worse. However, direct comparison may be misleading because different datasets are used, and accuracy also depends on the type of the entities in question. It is difficult to compare the work described here to that of Section 3.2.1, which attempts to disambiguate all types of concepts. Named entity disambiguation approaches have distinct and varying goals among themselves, and named entities have intrinsic properties that are not shared by common nouns.

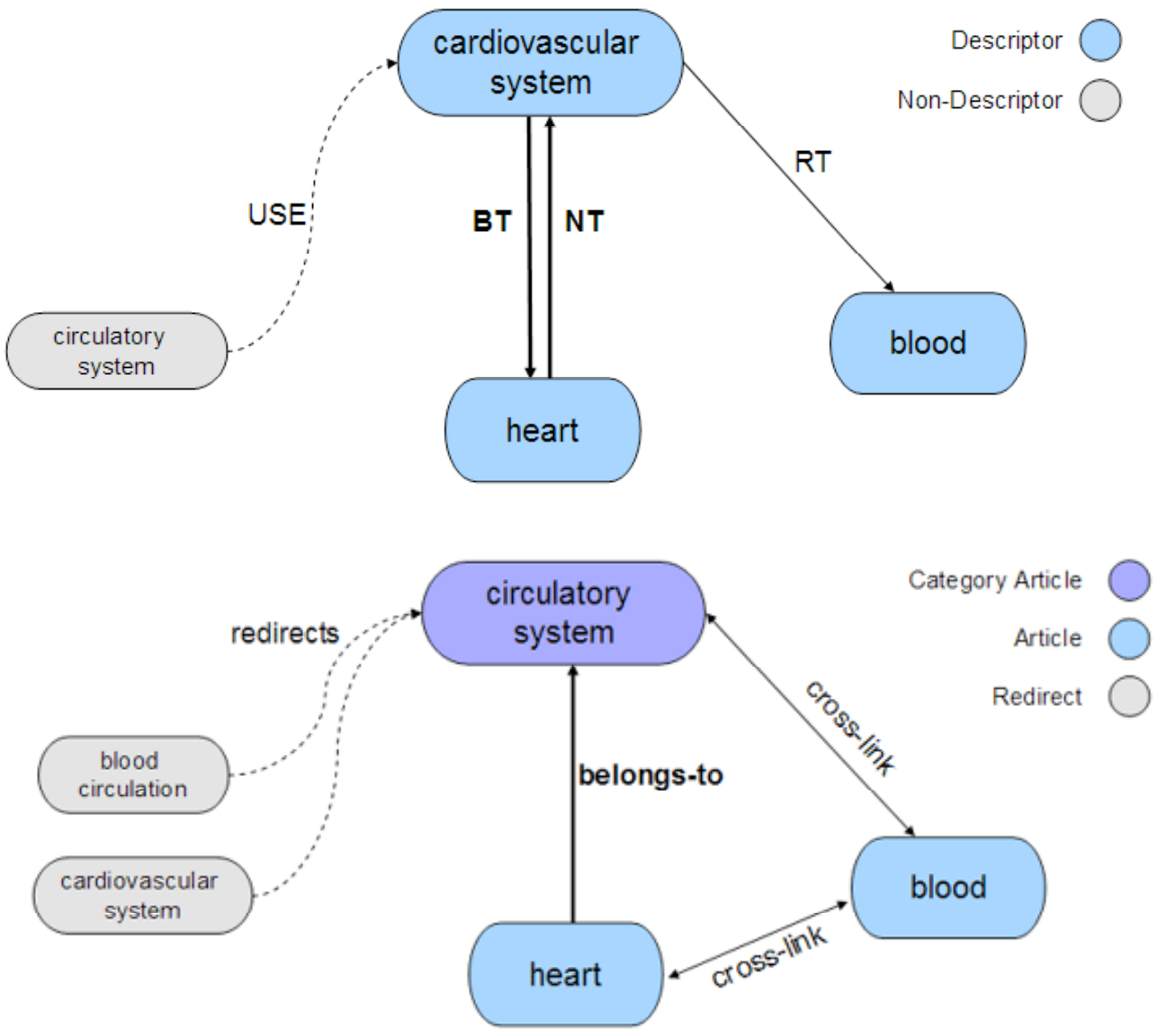

Figure 4. Comparison of organization structure in Agrovoc and Wikipedia.

*3.2.3. Disambiguating thesaurus and ontology terms.* Wikipedia's category and link structure contains the same kind of information as a domain-specific thesaurus, as illustrated by Figure 4, which compares it to the agricultural thesaurus Agrovoc [1995]. Section 3.1.2 used Wikipedia as an independent knowledge base, but it can also be used to extend and improve existing resources. For example, if it were known that *cardiovascular system* and *circulatory system* in Figure 4 refer to the same concept, the synonym *blood circulation* could be added to Agrovoc. The major problem is to establish a mapping between Wikipedia and other resources, disambiguating multiple mappings.

Ruiz-Casado et al. [2005a] map Wikipedia articles to WordNet. They work with the Simple Wikipedia mentioned earlier. WordNet synsets cluster word senses so that homonyms can be identified. If a Wikipedia article matches several WordNet synsets, the appropriate one is chosen by computing the similarity between the Wikipedia entry word-bag and the WordNet synset gloss. Dot product similarity of stemmed word vectors achieves 84% accuracy. The problem is that as Wikipedia grows, so does ambiguity. Even the Simple Wikipedia contains the article *Cats* (*musical*), though WordNet does not. The mapping technique must be able to deal with missing items as well as polysemy in both resources.

Overell and Rüger [2006] disambiguate place names mentioned in Wikipedia to locations in gazetteers. Instead of semantic similarity they develop geographically-based disambiguation methods. One uses geographical coordinates from the gazetteer to place a minimum bounding box around the location being disambiguated and other place names mentioned in the same context. Another analyzes the place name's referent; for example, if the surface form *Ontario* is mapped to *Ontario, Canada*, then *London, Ontario* can be mapped to *London, Canada*. Best results were achieved by combining the minimum bounding box method with "importance," measured by population size. An F-measure of 80% was achieved on a test set with 1,700 locations and 12,275 non-locations.

Overell and Rüger [2007] extend this approach by creating a co-occurrence model for each place name. They map place names to Wikipedia articles, collect their redirects as synonyms, and gather the anchor text of links to these articles. This yields different ways of referring to the same place, e.g., {*Londinium* → *London*} and {*London*, *UK* → *London*}. Next they collect evidence from Wikipedia articles: geographical coordinates, and location names in subordinate categories. They also mine Placeopedia, a mash-up website that connects Wikipedia with Google Maps. Together, these techniques recognize 75% of place names and map them to geographical locations with an accuracy of between 78–90%.

Milne et al. [2007] investigate whether domain-specific thesauri can be obtained from Wikipedia for use in natural language applications within restricted domains, comparing it with Agrovoc, a manually built agricultural thesaurus. On the positive side, Wikipedia article titles cover the majority of Agrovoc terms that were chosen by professional indexers as index terms for an agricultural corpus, and its redirects correspond closely with Agrovoc's synonymy relation. However, neither category relations nor (mutual) hyperlinks between articles correspond well with Agrovoc's taxonomic relations. Instead of extracting new domain-specific thesauri from Wikipedia they examine how existing ones can be improved, using Agrovoc as a case study [Medelyan and Milne 2008]. Given an Agrovoc descriptor, they collect semantically related terms from the Agrovoc hierarchy as context terms and map each one to the Wikipedia articles whose conditional probability (as explained in Section 3.2.1) is greatest. Then they compute the semantic similarity of each candidate mapping to this set of context articles. Manual evaluation of a subset with 400 mappings shows an average accuracy of 92%.

Medelyan and Legg [2008] map 50,000 terms from the Cyc ontology to Wikipedia articles using the disambiguation approach proposed by Medelyan and Milne [2008]. For each Cyc term, its surrounding ontology is used to gather a context for disambiguation, using the taxonomic relations *#$genls*, *#$isa* and some specific relations like *#$countryOfCity* and *#$conceptuallyRelated*. The most common Wikipedia article for each context term is identified and compared with all candidates for a mapping. A further test is applied when several Cyc terms map to the same Wikipedia article—reverse disambiguation. First, mappings that score less than 30% of the highest score are eliminated. A common-sense test is applied to the remainder based on Cyc's ontological knowledge regarding disjointness between classes. Evaluation shows that the mapping algorithm compares well with human performance.

In summary, despite the fact that there is still far less research on word sense disambiguation using Wikipedia than using WordNet, significant advances have been made. Over the last two years the accuracy of mapping documents to relevant Wikipedia articles has improved by one third [Milne and Witten 2008b]. Other researchers (such as Wang et al. [2007a]) use word sense disambiguation as a part of an application but do not provide any intrinsic evaluation. Furthermore, for fair comparison the same version of Wikipedia and the same training and test set should be used, as has been done for WordNet by SIGLEX (Section 3.2.1).

It is difficult to compare concept mapping techniques with one another, because each method concentrates on a different knowledge resource: WordNet, gazetteers, domain-specific thesauruses and Cyc. We look forward to more competition in these tasks, and to interesting applications of the resulting mappings.

A similar picture is observed in named entity extraction research, where each research group concentrates on different types of entity, e.g. persons or places. Here, extrinsic evaluations may be helpful—performance on a particular task like question answering before and after integration with Wikipedia. The next section describes an extrinsic evaluation of Wikipedia for co-reference resolution and compares the results with WordNet.

### 3.3 Co-reference resolution

Natural language understanding tasks such as textual entailment and question answering involve identifying which text entities refer to the same concept. Unlike word sense disambiguation, it is not necessary to determine the actual meaning of these entities, but merely identify their connection. Consider the following example from Wikipedia's article on New Zealand:

> **Elizabeth II**, as **the Queen** of New Zealand, is the Head of State and, in **her** absence, is represented by a non-partisan Governor-General. **The Queen** "reigns but does not rule." **She** has no real political influence, and **her** position is essentially symbolic. [emphasis added]

Without knowing that *Elizabeth II* and *the Queen* refer to the same entity, which can be referred to by the pronouns *she* and *her*, the information that can be inferred from this paragraph is limited. To resolve the highlighted co-referent expressions requires linguistic knowledge and world knowledge—that *Elizabeth II* is *the Queen*, and female. Current methods often derive semantic relations from WordNet or mine large corpora using lexical Hearst style patterns such as *X is a Y* and *Y such as X* [Hearst 1992]. The task can be modeled as a binary classification problem—to determine, for each pair of entities, whether they co-refer or not—and addressed using machine learning techniques, with features such as whether they are semantically related, the distance between them, agreement in number and gender.

The use of Wikipedia has been explored in two ways. Ponzetto and Strube [2006, 2007c] analyze its hyperlink structure and text to extract semantic features; whereas Yang and Su [2007] use it as a large semi-structured corpus for mining lexical patterns. Both use test data from the Message Understanding Conference organized by NIST.

Ponzetto and Strube's [2006, 2007c] goal is to show that Wikipedia can be used as a fully-fledged lexical and encyclopedic resource, comparable to WordNet but far more extensive. While their work on semantic relatedness (Section 3.1) evaluates Wikipedia intrinsically, co-reference is evaluated extrinsically to demonstrate Wikipedia's utility. As a baseline they re-implement Soon et al.'s [2001] method with a set of standard features, such as whether the two entities share the same grammatical features or WordNet class. Features mined from WordNet and Wikipedia are evaluated separately. The WordNet features for two given terms A (*Elizabeth II*) and B (*the Queen*) are:

- the highest similarity score from all synset pairs to which A and B belong,
- the average similarity score.

The Wikipedia analogue to these two features,

- the highest similarity score from all Wikipedia categories of A and B,
- the average similarity score,

is augmented by further features:

- Does the first paragraph of the Wikipedia article describing A mention B?
- Does any hyperlink in A's article target B?
- Does the list of categories for A's article contain B?
- What is the overlap between the first paragraphs of the articles for A and B?

The similarity and relatedness scores are computed using various metrics. Feature selection is applied during training to remove irrelevant features for each scenario. Results are presented in Table 3, which we will discuss shortly.

| | | NWIRE | | | BNEWS | | |
|---|---|---|---|---|---|---|---|
| | | R | P | F | R | P | F |
| Ponzetto and Strube [2006, 2007c] | baseline | 56.3 | 86.7 | 68.3 | 50.5 | 82.0 | 62. |
| | +WordNet | 62.4 | 81.4 | 70.7 | 59.1 | 82.4 | 68. |
| | +Wikipedia | 60.7 | 81.8 | 69.7 | 58.3 | 81.9 | 68. |
| Yang and Su [2007] | baseline | 54.5 | 80.3 | 64.9 | 52.7 | 75.3 | 62. |
| | +sem. related. | 57.4 | 80.8 | 67.1 | 54.0 | 74.7 | 62. |

Table 3. Performance comparison of two independent techniques on the same datasets.

Yang and Su [2007] utilize Wikipedia in a different way, assessing semantic relatedness between two entities by analyzing their co-occurrence patterns in Wikipedia. (Pattern matching using the Wikipedia corpus is practiced extensively in information extraction; see Section 5). The patterns are evaluated based on positive instances in the training data that serve as seeds. For example, given the pair of co-referents *Bill Clinton* and *president*, and Wikipedia sentences like *Bill Clinton is elected President of the United States* and *The US president, Mr Bill Clinton*; the patterns [*X is elected Y*] and [*Y, Mr X*] are extracted. Sometimes patterns occur in structured parts of Wikipedia like lists and infoboxes—for example, the bar symbol is the pattern in *United States | Washington, D.C.* An accuracy measure is used to eliminate patterns that are frequently associated with both negative and positive pairs. Yang and Su [2007] found that using the 100 most accurate patterns as features did not improve performance over the baseline. However, adding a single feature representing semantic relatedness between the two entities did improve results. Yang and Su use mined patterns to assess relatedness by multiplying together two measures of reliability: the strength of association between each positive seed pair and the pointwise mutual information between the entities occurring with the pattern and by themselves.

Table 3 shows the results that both sets of authors report for co-reference resolution. They use the same baseline, but the implementation was evidently slightly different, for Ponzetto and Strube's yielded a slightly improved F-measure. Ponzetto and Strube's results when features were added from WordNet and Wikipedia are remarkably similar, with no statistical difference between them. These features decrease precision over the baseline on NWIRE by 5 points but increase recall on both datasets, yielding a significant overall gain (1.5 to 2 points on NWIRE and 6 points on BNEWS). Yang and Su improve the F-measure on NWIRE and recall on BNEWS by 2 points. Overall, it seems that Ponzetto and Strube's technique performs slightly better.

These co-reference resolution systems are quite complex, which may explain why no other methods have been described in the literature. We expect further developments in this area.

## 3.4 Multilingual alignment

In 2006, five years after its inception, Wikipedia contained 100,000 articles for eight different languages. The closest precedent to this unique multilingual resource is the commercial EuroWordNet that unifies seven different languages but covers a far smaller set of concepts—8,000 to 44,000, depending on the language [Vossen et al. 1997]. Of course, multilingual vocabularies and aligned corpora benefit any application that involves machine translation.

Adafre and de Rijke [2006] began by generating parallel corpora in order to identify similar sentences—those whose information overlaps significantly—in English and Dutch. First they used a machine translation tool to translate Wikipedia articles and compared the result with the corresponding manually written articles in that language. Next they generated a bilingual lexicon from links between articles on the same topic in different languages, and determined sentence similarity by the number of shared lexicon entries. They evaluated these two techniques manually on 30 randomly chosen Dutch and English Wikipedia articles. Both identified rather a small number of correct sentence alignments; the machine translation had lower accuracy but higher coverage than the lexicon approach. The authors ascribed the poor performance to the small size of the Dutch version but were optimistic about Wikipedia's potential.

Ferrández et al. [2007] use Wikipedia for cross-language question answering (Section 4.3 covers research on monolingual question answering). They identify named entities in the query, link them to Wikipedia article titles, and derive equivalent translations in the target language. Wikipedia's exceptional coverage of named entities (Section 3.2.2) counters the main problem of cross-language question answering: low coverage of the vocabulary that links questions to documents in other languages. For example, the question *In which town in Zeeland did Jan Toorop spend several weeks every year between 1903 and 1924?* mentions the entities *Zeeland* and *Jan Toorop*, neither of which occurs in EuroWordNet. In an initial version of the system using that resource, *Zeeland* remains unchanged and *the phrase Jan Toorop* is translated to *Enero Toorop* because *Jan* is erroneously interpreted as *January*. With Wikipedia as a reference, the translation is correct: *¿En qué ciudad de Zelanda pasaba varias semanas al año Jan Toorop entre 1903 y 1924?* Wikipedia's coverage allowed Ferrández et al. to increase the number of correctly answered questions by 20%.

Erdmann et al. [2008] show that simply following language links in Wikipedia is insufficient for a high-coverage bilingual dictionary. They develop heuristics based on Wikipedia's link structure that extract significantly more translation pairs, and evaluate them on a manually created test set containing terms of different frequency. Given a Wikipedia article that is available in another language—the target article—they augment the translated article name with redirects and the anchor text used to refer to the article. Redirects are weighted by the proportion of links to the target article (including redirects) that use this particular redirect. Anchors are weighted similarly, by expressing the number of links that use this particular anchor text as a proportion of the total number of incoming links to the article. The resulting dictionary contains all translation pairs whose weight exceeds a certain threshold. This achieves significantly better results than a standard dictionary creation approach using parallel corpora.

We know of only a few disjointed examples that draw on Wikipedia as multilingual corpus. Sections 4.2 and 4.3 discuss a few more, but we expect further advances. The data mined by these algorithms will continue to grow: between 2001 and 2006 the proportion of active Wikipedians contributing to non-English Wikipedias rose from 22% to 55%.[14] With this increasing diversity, it will not be surprising to see Wikipedia become a prominent resource for machine translation, multi-lingual information retrieval, and other such tasks.

This section has described several natural language processing techniques that benefit from Wikipedia's immense potential as a repository of linguistic knowledge. Well-defined tasks such as determining semantic relatedness and word sense disambiguation have been significantly improved within just a few years. Many other natural language tasks have hardly been touched by Wikipedia research—automatic text summarization, text generation, text segmentation, machine translation, parsing. Wikipedia's rapidly growing potential as a multilingual knowledge source has been already explored in cross-language retrieval (Section 4.2), and we expect accelerating enthusiasm from machine translation researchers. The Simple Wikipedia, where concept definitions are expressed using basic English phrases and sentences, has unexplored potential for further linguistic work such as syntactic parsing and language learning tools.

## 4. INFORMATION RETRIEVAL

Wikipedia is already one of the most popular web sites for locating information. Here we ask how it can be used to make information easier to obtain from elsewhere—how to apply it to organize and locate other resources.

Given its applications for natural language processing (Section 3), it is not surprising to see Wikipedia leveraged to gain a deeper understanding of both queries and documents, and improve how they are matched to each other. Section 4.1 describes how it has been used to expand queries to allow them to return more relevant documents, while Section 4.2 describes experiments in cross-language retrieval. Wikipedia has also been used to retrieve specific portions of documents, such as answers to questions (Section 4.3) or important entities (Section 4.4).

[14] These figures were gathered from *http://stats.wikimedia.org*. For some reason, statistics for the English Wikipedia are not updated beyond October 2006.

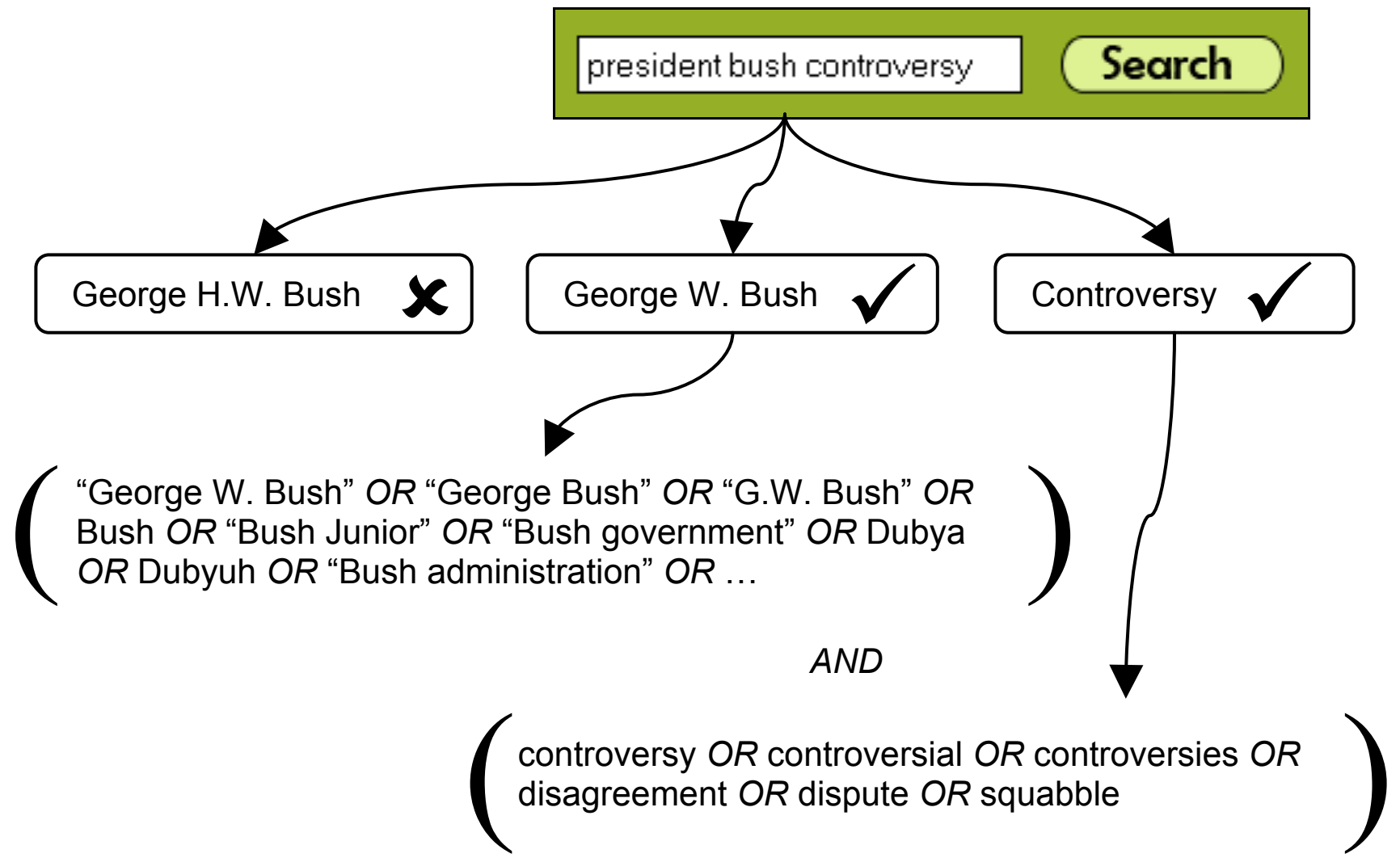

Figure 5. Using Wikipedia to recognize and expand query topics.

The same Wikipedia-derived understanding has been used to automatically organize documents into helpful groups. Section 4.5 shows how Wikipedia has been applied to document classification, where documents are categorized under broad headings like *Sport* and *Technology*. To a lesser extent it has also been used to determine the main topics that documents discuss, so that they can be organized under specific keyphrases (Section 4.6).

## 4.1 Query expansion

Query expansion aims to improve queries by adding terms and phrases, such as synonyms, alternative spellings, and closely related concepts. Such reformulations can be performed automatically—without the user's input—or interactively—where the system suggests possible modifications.

Gregorowicz and Kramer [2006] were among the first to see in Wikipedia a solution to "the problem of variable terminology." Their goal is to construct a comprehensive term-concept map that facilitates concept-based information retrieval by resolving synonyms in a systematic way. For this they use Wikipedia articles as concepts, and establish synonyms via redirects and homonyms via disambiguation pages. This produces a network of 2M concepts linked to 3M terms—a vast and impressive resource compared to WordNet's 115,000 synsets created from 150,000 words.

Milne et al. [2007] use Wikipedia to provide both forms of expansion in their knowledge-based search engine Koru.[15] They first obtain a subset of Wikipedia articles that are relevant for a particular document collection, and use the links between these to build a corpus-specific thesaurus. Given a query they map its phrases to topics in this thesaurus. Figure 5 illustrates how the query *president bush controversy* is mapped to potentially relevant thesaurus topics (i.e., Wikipedia articles) *George H.W. Bush*, *George W. Bush* and *Controversy*. *President Bush* is initially disambiguated to the younger of the two, because he occurs most often in the document set. This can be corrected manually. The redirects from his article and that of *Controversy* are then mined for synonyms and alternative spellings, such as *Dubya* and *disagreement*, and quotes are added around multi-word phrases (such as *Bush administration*). This results in a complex Boolean query such as an expert librarian might issue. The knowledge base was capable of recognizing and lending assistance to 95% of the queries issued to it. Evaluation over the TREC HARD Track [Allan 2005] shows that the expanded queries are significantly better than the original ones in terms of overall F-measure.

Milne et al. also provided interactive query expansion by using the detected query topics as starting points for browsing the Wikipedia-derived thesaurus. For example, *George Bush* provides a starting point for locating related topics such as *Dick Cheney*, *Terrorism*, and *President of the United States*. The evaluation of such exploratory search provided little evidence that it assisted users. Despite this, the authors argue that Wikipedia should be an effective base for this task, due to its extensive coverage and inter-linking. This is yet to be proven, and we know of no other examples of exploratory searching with Wikipedia.

Li et al. [2007a] also use Wikipedia to expand queries, but focus on those that traditional approaches fail to improve. The standard method of pseudo-relevance feedback works by feeding terms from the highest ranked documents back into the query [Ruthven and Lalmas 2003]. This works well in general, so most state-of-the-art approaches are variants of this idea. Unfortunately it makes bad queries even worse, because it relies on at least the top few documents being relevant. Li et al. avoid this by using Wikipedia as an external corpus to obtain additional query terms. They issue the query on Wikipedia to retrieve relevant articles, use these articles' categories to group them, and rank articles so that those in the largest groups appear more prominently. Forty terms are then picked from the top 20 articles (it is unclear how they are selected) and added to the original query. When tested on queries from TREC's 2005 Robust track [Allan 2005], this improved those queries for which traditional pseudo-relevance feedback performs most poorly. However, it did not achieve the state of the art in general. The authors attribute this to differences in language and context between Wikipedia and the (dated) news articles used for evaluation, which render many added terms irrelevant.

Where the previous two systems departed from traditional bag-of-words relevance feedback, Egozi et al. [2008] instead aim to augment it. Their system uses ESA (Section 3.1) to represent documents and queries as vectors of their most relevant Wikipedia articles. Comparison of document vectors to the query vector results in concept-based relevance scores, which are combined with those given by state-of-the-art retrieval systems such as Xapian and Okapi. Additionally, both concept-based and bag-of-words scores are computed by segmenting documents into overlapping 50-word subsections (a common strategy), and a document's total score is the sum of the score obtained from its best section and its overall content. One drawback is ESA's tendency to provide features (Wikipedia articles) that are only peripherally related to queries. The query *law enforcement, dogs,* for example, yields not just *police dog* and *cruelty to animals* but also *contract* and *Louisiana.* To counter this documents are ranked according to their bag-of-words scores and the highest and lowest ranking documents are used to provide positive and negative examples for feature selection. When used to augment the four top performing systems from the TREC-8 competition [Voorhees and Harman 2000] this improved Mean Average Precision by 4–15% depending on the system being augmented.

Wikipedia seems well suited to query expansion. Bag-of-words approaches stand to benefit from its knowledge of what the words mean and how they relate to each other. Concept-based approaches that draw on traditional knowledge bases could profit just as much from its unmatched breadth. We expect widespread application of Wikipedia in the future for both automatic query expansion and exploratory searching, It will improve existing techniques and support entirely new ones.

### 4.2 Multilingual retrieval

Multilingual or cross-language information retrieval involves seeking relevant documents that are written in a language different to the query. Wikipedia has clear application here. Although its language versions grow at different rates and cover different topics, they are carefully interwoven. For example, the English article on *Search engines* is linked to the German *Suchmaschine*, the French *Moteur de recherché*, and more than 40 other translations. These links constitute a comprehensive and rapidly growing cross-lingual dictionary of topics and terms. Wikipedia is ideal for translating emerging named entities and topics, such as people and technologies—exactly the items that traditional multilingual resources (dictionaries) struggle with. Surprisingly, we failed to locate any papers that use Wikipedia's cross-language links directly to translate query topics.

[15] Demo at *http://www.nzdl.org/koru*

Potthast et al. [2008] jump directly to a more sophisticated solution that uses Wikipedia to generate a multilingual retrieval model. This is a generalization of traditional monolingual retrieval models like vector spaces and latent semantic analysis that assess similarity between documents and text fragments. Multilingual and cross-language models are capable of identifying similar documents even when written in different languages. Potthast et al. use ESA (Section 3.1) as the starting point for a new model called Cross-language Explicit Semantic Analysis (CL-ESA). Their approach hypothesizes that the relevant concepts identified by ESA are essentially language independent, so long as the concepts are sufficiently described in different languages. If there were sufficient overlap between the English and German Wikipedias, for example, one would get roughly the same list of concepts (and in the same order) from ESA regardless of whether the document being represented, or the concept space it was projected onto, was in English or German. This makes the language of documents and concept spaces largely irrelevant, so that documents in different languages can be compared without explicit translation.

To evaluate this idea, Potthast et al. conducted several experiments with a bilingual (German–English) set of 3,000 documents. One test was to use articles in one language as queries to retrieve their direct translation in the other language. When CL-ESA was used to rank all English documents by their similarity to German ones, the explicit translation of the document was consistently ranked highly—it was top 91% of the time, and in the top ten 99% of the time. Another test was to use an English document as a query for the English document set, and its translation as a query for the German one. The two result sets had an average correlation of 72%. These results were obtained with a dimensionality of $10^5$; that is, 100,000 bilingual concepts were used to generate the concept spaces. Today, only German and English Wikipedias have this degree of overlap. Results degrade as fewer concepts are used; 1,000–10,000 concepts were deemed sufficient for reasonable retrieval performance. At the time this made CL-ESA capable of pairing English with German, French, Polish, Japanese, and Dutch.

As with the work described in Section 3.4, the results will only become better and more broadly applicable. Even if the algorithm itself does not improve, Wikipedia's continued growth will allow this and other techniques to provide more accurate responses and be applied to more and more languages over time.

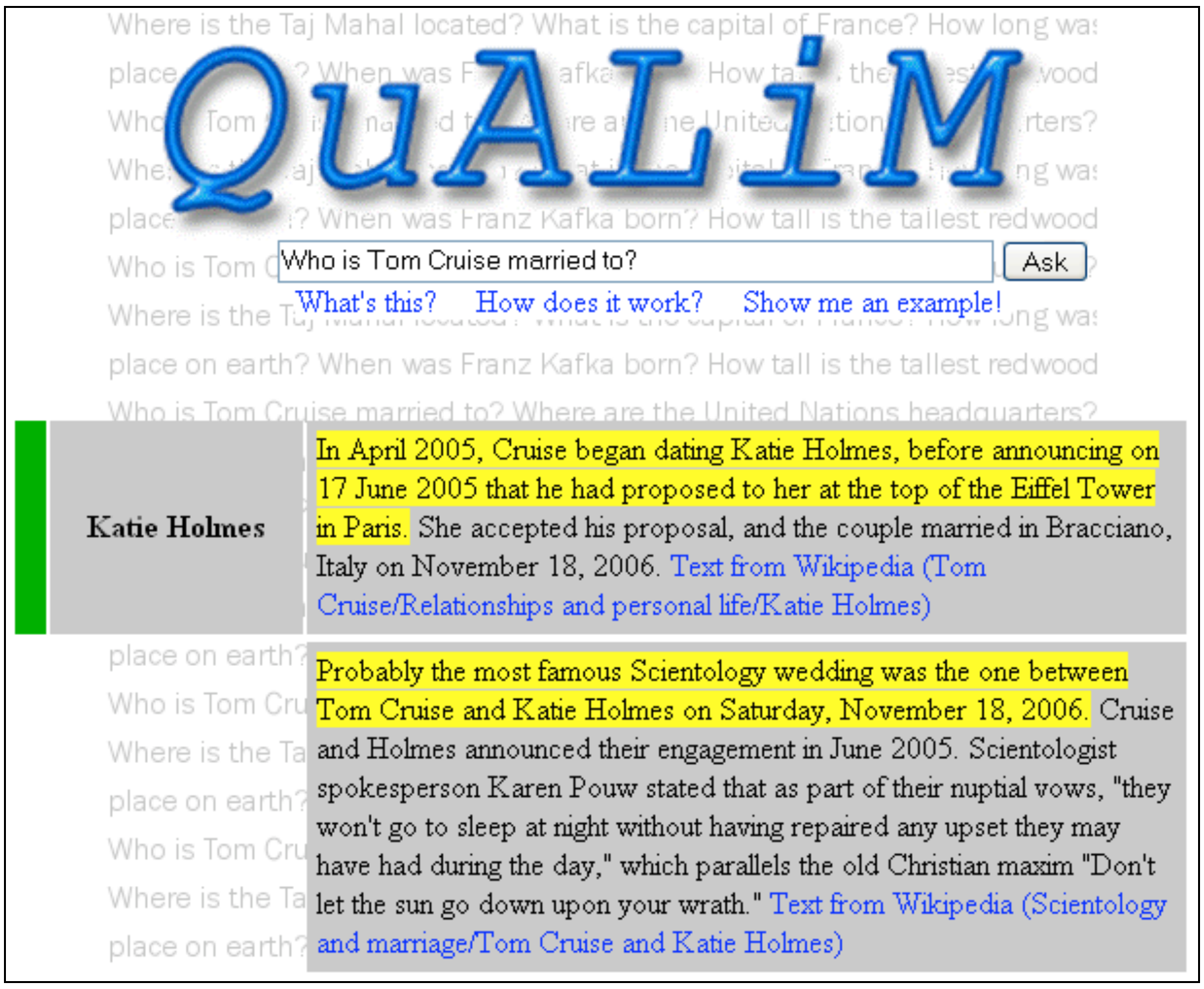

Figure 6. The QuALiM system, using Wikipedia to answer *Who is Tom Cruise married to?*

### 4.3 Question answering

Question answering aims to return specific answers to questions, rather than entire documents. Approaches range from extracting the most relevant sentences or sections from documents, to ensuring that they are in the correct form to constitute an answer, to constructing answers on the fly. Wikipedia, a broad corpus filled with numerous facts, is a promising source of answers. A simple but well-known example is the fact that Google queries prefixed with *define*, and Ask.com queries starting with *What is…* or *Who is…*, often return the first sentence from the relevant Wikipedia article.

Kaisser's [2008] QuALiM system, illustrated in Figure 6, is more sophisticated.[16] When asked a question (such as *Who is Tom Cruise married to?*) it mines Wikipedia not only for relevant articles, but for the sentences and paragraphs that contain the answer. It also provides the exact entity that answers the question—e.g. *Katie Holmes*. Interestingly, this entity is not mined from Wikipedia but obtained by analyzing results from web search engines. Questions are parsed to identify the expected class of the answer (in this case, a person) and construct valid queries (e.g. *Tom Cruise is married to* or *Tom Cruise's wife*).

The responses are parsed to identify entities of the appropriate type. Wikipedia is only used to supply the supporting sentences and paragraphs.

The TREC conferences are a prominent forum for investigating question answering.[17] The question-answering track provides ground truth for experiments with a corpus from which answers to questions have been extracted manually. The 2004 track saw two early uses of Wikipedia for question answering: Lita et al. [2004] and Ahn et al. [2004]. The former does not answer questions *per se*; instead it investigates whether different resources provide the answers without attempting to extract them. Wikipedia's coverage of answers was 10 percentage points greater than WordNet's, and about 30 points greater than other resources including Google *define* queries and gazetteers such as the CIA *World Fact Book*.

Ahn et al. [2004] seem to be the first to provide explicit answers from Wikipedia. They identify the question's topic—*Tom Cruise* in our example—and locate the relevant article. They then identify the expected type of the answer—in this case, another person (his wife)—and scan the article for matching entities. These are ranked by both prior answer confidence (probability that they answer any question at all) and posterior confidence (probability that they answer the question at hand). Prior confidence is given by the entity's position in the article, the most important facts are covered first. Posterior confidence is given by the Jaccard similarity of the question to the sentence surrounding the entity. Wikipedia is used as one stream among many from which to extract answers. Unfortunately the experiments do not tease out its individual contribution. Overall, however, they did not improve upon their previous work.

The CLEF series of conferences is another popular forum for investigating question answering.[18] Corpora and tasks in many different languages are provided for monolingual and cross-language work, one source of documents being a cross-language crawl of Wikipedia. Most competition entries extract answers from Wikipedia, but are not covered here because they do not take advantage of its unique properties.

Buscaldi and Rosso [2007a] use Wikipedia to augment their question answering system. They left unchanged the way in which this system extracts answers except for an additional step where Wikipedia is consulted to verify the results. They index four different views of Wikipedia—titles, full text, first sections (definitions), and the categories that articles belong to—and search them differently depending on the question type. Answers to definition questions (e.g., *Who is Nelson Mandela?*) are verified by seeking articles whose title contains the corresponding entity and whose first section contains the proposed answer. If the question requires a name (e.g., *Who is the President of the United States?*) the process is reversed: candidate answers (*Bill Clinton, George Bush*) are sought in the title field and query constraints (*President*, *United States*) in the definition. In either case, if at least one relevant article is returned the answer is verified. This yielded an improvement of 4.5% over the original system across all question types. Ferrández et al. [2007] also use Wikipedia's structure to answer questions, but focus on cross-lingual tasks (see Section 3.4).

[16] Demo at *http://demos.inf.ed.ac.uk:8080/qualim/*
[17] *http://trec.nist.gov/*
[18] The homepage for the CLEF series of conferences is at *http://www.clef-campaign.org/*

As well as using Wikipedia as a corpus for standard question answering, CLEF has a track specifically designed to assist Wikipedia's contributors. Given a source article, the aim is to extract new snippets of information from related articles that should be incorporated into it. Jijkoun and de Rijke [2006] conclude that the task is difficult but possible, so long as the results are used in a supervised fashion. The best of seven participating teams added an average of 3.4 "perfect" (important and novel) snippets to each English article, with a precision of 36%. Buscaldi and Rosso [2007b], one of the contributing entries,[19] search Wikipedia for articles containing the target article's title. They extract snippets, rank them according to their similarity to the original article using the bag-of-words model, and discard those that are redundant (too similar) or irrelevant (not similar enough). On English data this yields 2.7 perfect snippets per topic, with a precision of 29%. On Spanish data it obtains 1.8 snippets with 23% precision.

Finally, Higashinaka et al. [2007] extract questions, answers and even hints from Wikipedia to automatically generate "*Who am I?*" quizzes. The first two tasks are simple because the question is always the same and the answer is always a person. The challenging part is extracting hints (essentially, facts about the person) and ranking them so that they progress from vague to specific. They used machine learning based on biographical Wikipedia articles whose facts have been manually ranked.

Overall, research on question answering tends to treat Wikipedia as just another plain-text corpus. Few researchers capitalize on its unique structural properties (categories, links, etc.) or the explicit semantics it provides. For example, standard word-based similarity measures continue to be applied, even though concept-based measures such as ESA have been proven more effective. There is little overlap between this work and research on information extraction from Wikipedia, and no use of Wikipedia-derived ontologies, or its infoboxes (Section 5). This reflects an overall philosophy of crawling the entire web for answers, requiring techniques that generalize to any textual resource.

[19] We were unable to locate papers describing the others.

### 4.4 Entity ranking

It is often expedient to return entities in response to a query rather than full documents as in classical retrieval. This resembles question answering, and often fulfils the same purpose—for example, the query *countries where I can pay in euros* could be answered by a list of relevant countries. For other queries, however, entity ranking does not provide answers but instead generates a list of pertinent topics. For example, as well as *Google*, *Yahoo*, and *Microsoft Live* the query *search engines* would also return *PageRank* and *World Wide Web.* The literature seems to use 'entity' and 'named entity' interchangeably, making it unclear whether concepts such as *information retrieval* and *full text search* would also be valid results.

Section 5.2 demonstrates that Wikipedia offers an exceptionally large pool of manually-defined entities, which can be typed (as people, places, events, etc.) fairly accurately. The entity ranking track of the Initiative for Evaluation of XML Retrieval (INEX) compares different methods for entity ranking by how well they are able to return relevant Wikipedia entities in response to queries [de Vries et al. 2007]. Zaragoza et al. [2007] use Wikipedia as a dataset for comparing two approaches to entity ranking: entity containment graphs and web search based methods—their results are not described here because they do not relate directly to Wikipedia. However, they have developed a version of Wikipedia in which named entities have been automatically annotated, and are sharing it so that others can investigate different approaches to named entity ranking.[20]

Wikipedia provides a wealth of information about the entities it contains, which can improve ranking. Vercoustre et al. [2008] combine traditional search with Wikipedia-specific features. They rank articles (which they assume are synonymous with entities) by combining the score provided by a search engine (Zettair) with features mined from categories and inter-article links. The latter provide a simplified PageRank for entities and the former a similarity score for how they relate to each other. The resulting precision almost doubles that of the search engine alone. Vercoustre et al. were the only competitors for the INEX entity-ranking track we were able to locate,[21] and it seems that Wikipedia's ability to improve entity ranking has yet to be evaluated against more sophisticated baselines. Moreover, the features that they derive from Wikipedia are only used to rank entities in general, not by their significance for the query. Regardless, entity ranking will no doubt receive more attention as the INEX competition grows and others use Zaragossa et al.'s dataset.

[20] The annotated version of Wikipedia is at *http://www.yr-bcn.es/semanticWikipedia*
[21] It began in 2007 and the Proceedings are yet to be published.

| | Micro BEP | Macro BEP |
|---|---|---|
| Baseline (from Gabrilovich and Markovitch [2006]) | 87.7 | 60.2 |
| Gabrilovich and Markovitch [2006] | 88.0 | 61.4 |
| Wang et al. [2007a] | 91.2 | 63.1 |

Table 4. Performance of text categorization over the Reuters-21578 collection.

The knowledge that Wikipedia provides about entities can also be used to organize them. This has not yet been thoroughly investigated, the only example being Yang et al.'s [2007] use of Wikipedia articles and WikiBooks to organize entities into hierarchical topic maps. They search for the most relevant article and book for a query and simply strip away the text to leave lists of links—which again they assume to be entities—under the headings in which they were found. This is both an entity ranking method and a tool for generating domain-specific taxonomies, but has not been evaluated as either.

Entity ranking is a young field and research is sparse. Overall, it seems that researchers view entity ranking over Wikipedia—where entities and the information pertaining to them are clearly demarcated—as the low-hanging fruit. It will be interesting to see what challenges are involved in generalizing this research to utilize other resources.

### 4.5 Text categorization

Text categorization (or classification) organizes documents into labeled groups, where labels come from a pool of pre-determined categories. The traditional approach is to represent documents by the words they contain, and use training documents to identify words and phrases that are indicative of each category label. Wikipedia allows categorization techniques to draw on background knowledge about the concepts the words represent. As Gabrilovich and Markovitch [2006] note, traditional approaches are brittle and break down when documents discuss similar topics in different terms—as when one talks of *Wal-Mart* and the other of *department stores*. They cannot make the necessary connections because they lack background knowledge about what the words mean. Wikipedia can fill the gap.

As a quick indication of its application, Table 4 compares Wikipedia-based approaches with state-of-the-art methods that only use information in the documents themselves. The figures were obtained on the Reuters-21578 collection, a set of news stories that have been manually assigned to categories. Results are presented as the breakeven point (BEP) where recall and precision are equal. The micro and macro columns correspond to how these are averaged: the former averages across documents, so that smaller categories are largely ignored; the latter averages by category. The first entry is a baseline provided by Gabrilovich and Markovitch, which is in line with state-of-the-art methods such as [Dumais et al. 1998]. The remaining three entries use additional information gleaned from Wikipedia and are described below. The gains may seem slight, but they represent the first improvement upon a performance plateau reached by established techniques that are now a decade old.

Gabrilovich and Markovitch [2006] observe that documents can be augmented with Wikipedia concepts without complex natural language processing. Both are in the same form—plain text—so standard similarity algorithms can be used to compare documents with potentially relevant articles. Thus documents can be represented as weighted lists of relevant concepts rather than bags of words. This should sound familiar—it is the predecessor of ESA (Sections 3.1, 4.1, 4.2), from the same authors. For each document, Gabrilovich and Markovitch generate a large set of features (articles) not just from the document as a whole, but also by considering each word, sentence, and paragraph independently. Training documents are then used to filter out the best of these features to augment the original bag of words. The number of links made to each article is used to identify and emphasize those that are most well known. This results in consistent improvement over previous classification techniques, particularly for short documents (which otherwise have few features) and categories with fewer training examples.

Wikipedia's ability to improve classification of short documents is confirmed by Banerjee et al. [2007], who cluster news articles under feed items such as those provided by Google News. They obtained relevant articles for each news story by issuing its title and short description (Google snippet) as separate queries to a Lucene index of Wikipedia. They were able to cluster documents under their original headings (each feed item organizes many similar stories) with 90% accuracy using only the titles and descriptions as input. However, this work treats Google's automatically clustered news stories as ground truth, and only compares the Wikipedia-based approach with a baseline of their own design.

Wang et al. [2007a] also use Wikipedia to improve document classification, but focus on mining it for terms and phrases to add to the bag of words representation. For each document they locate relevant Wikipedia articles by matching n-grams to article titles. They augment the document by crawling these articles for synonyms (redirects), hyponyms (parent categories) and associative concepts (inter-article links). Though the last yields many tenuous semantic relationships, these are refined by selecting linked articles that are closely related according to textual content or parent categories. As shown in Table 4, this results in the best overall performance.

As well as a source of background knowledge for improving classification techniques, Wikipedia can be used as a corpus for training and evaluating them. Almost all classification approaches are machine-learned, and thus require training examples. Wikipedia provides millions of them. Each association between an article and the categories to which it belongs can be considered as manually defined ground truth for how that article should be classified. Gleim et al. [2007], for example, use this to evaluate their techniques for categorizing web pages solely on structure rather than textual content. Admittedly, this is a well-established research area with well-known datasets, so it is unclear why another one is required. Table 4, for example, would be more informative if all of the researchers using Wikipedia for document classification had used standard datasets instead of creating their own.

| | Micro BEP | Macro BEP |
|---|---|---|
| Baseline (from Gabrilovich and Markovitch [2006]) | 87.7 | 60.2 |
| Gabrilovich and Markovitch [2006] | 88.0 | 61.4 |
| Wang et al. [2007a] | 91.2 | 63.1 |

Table 4. Performance of text categorization over the Reuters-21578 collection.

Two approaches that do not compete with the traditional bag-of-words (and will therefore be discussed only briefly) are Janik and Kochut [2007] and Minier et al. [2007]. The former is one of the few techniques that does not use machine learning for classification. Instead, miniature “ontologies”—rough networks of relevant concepts—are mined from Wikipedia for each document and category, and the most relevant category ontology to each document ontology is identified. The latter approach transforms the traditional document-term matrix by mapping it onto a gigantic term-concept matrix obtained from Wikipedia. PageRank is run over Wikipedia’s inter-article links in order to weight the derived concepts, and dimensionality reduction techniques (latent semantic analysis, kernel principle component analysis and kernel canonical correlation analysis) are used to reduce the representation to a manageable size. Minier et al. attribute the disappointing results (shown in Table 4) to differences in language usage between Wikipedia and the Reuters corpus used for evaluation. It should be noted that their Macro BEP (the highest in the table) may be misleading; their baseline achieves an even higher result, indicating that their experiment should not be compared to the other three.

Banerjee [2007] observed that document categorization is a problem where the goalposts shift regularly. Typical applications are organizing news stories or emails, which arrive in a stream where the topics being discussed constantly evolve. A categorization method trained today may not be particularly helpful next week. Instead of throwing away old classifiers, they show that inductive transfer allows old classifiers to influence new ones. This improves results and reduces the need for fresh training data. They find that classifiers that derive additional knowledge from Wikipedia are more effective at transferring this knowledge, which they attribute to Wikipedia's ability to provide background knowledge about the content of articles, making their representations more stable.

Dakka and Cucerzan [2008] and Bhole et al. [2007] perform the reverse operation. Instead of using Wikipedia to augment document categorization, they apply categorization techniques to Wikipedia. Their aim is to classify articles to detect the types (people, places, events, etc.) of the named entities they represent. This is discussed in Section 5.2 on named entity recognition. Also discussed elsewhere (Section 4.6) is Schönhofen [2006], who developed a topic indexing system but evaluated it as a document classifier.

Overall, the use of Wikipedia for text categorization is a flourishing research area. Many recent efforts have improved upon the previous state of the art; a plateau that had stood for almost a decade. Some of this success may be due to the amount of attention the problem has generated (at least ten papers in just three years), but more fundamentally it can be attributed to the way in which researchers are approaching the task. Just as we saw in Section 4.1, the greatest gains have come from drawing closely on and augmenting existing research, while thoroughly exploring the unique features that Wikipedia offers.

## 4.6 Topic indexing

Topic indexing is subtly different from text categorization. Both label documents so that they can be grouped sensibly and browsed efficiently, but in topic indexing labels are chosen from the topics the documents discuss rather than from a predetermined pool of categories. Topic labels are typically obtained from a domain-specific thesaurus—such as MESH [Lipscomb 2000] for the Medical domain—because general thesauri like WordNet and Roget are too small to provide sufficient detail. An alternative is to obtain labels from the documents themselves, but this is inconsistent and error-prone because topics are difficult to recognize and appear in different surface forms. Using Wikipedia as a source of labels sidesteps the onerous requirement for developing or obtaining relevant thesauri, since it is large and general enough to apply to all domains. It might not achieve the same depth as domain-specific thesauri, but tends to cover the topics that are used for indexing most often [Milne et al. 2006]. It is also more consistent than extracting terms from the documents themselves, since each concept in Wikipedia is represented by a single succinct manually chosen title. In addition to the labels themselves, Wikipedia provides many features about the concepts, such as how important and well known they are and how they relate to each other.

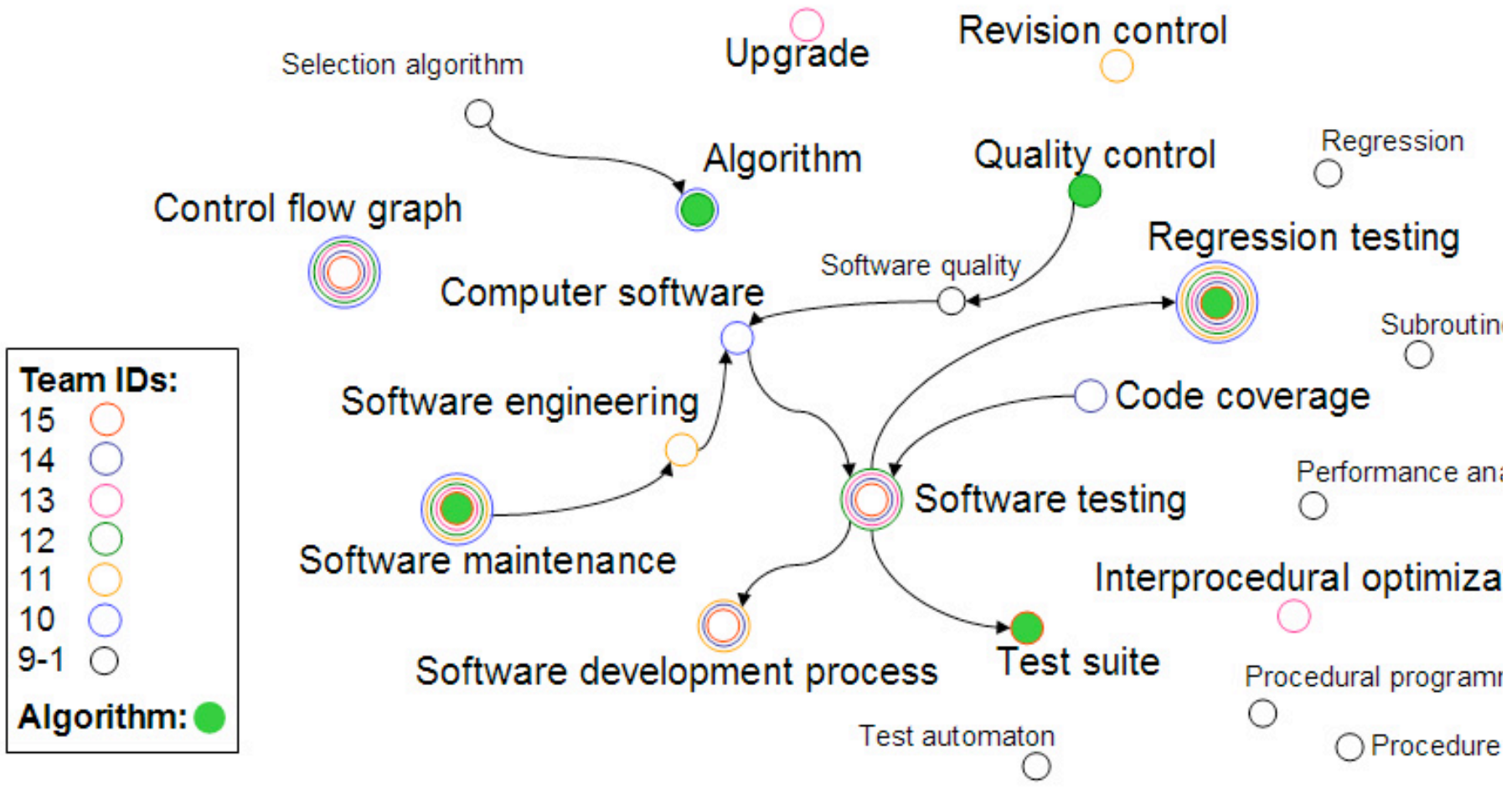

Figure 7. Topics assigned to a document entitled "A Safe, Efficient Regression Test Selection Technique" by human teams (outlined circles) and the new algorithm (filled circles).

Medelyan et al. [2008] use Wikipedia as a controlled vocabulary and apply *wikification* (Section 3.2.1) to identify the topics that documents mention. For each candidate topic they identify several features, including classical ones such as how often topics are mentioned, and two Wikipedia-specific ones. One is *node degree*: the extent to which each candidate topic (article) is linked to the other topics detected in the document. The other is *keyphraseness*: the extent to which the topics are used as links in Wikipedia. They use a supervised approach that learns the typical distributions of these features from a manually tagged corpus [Frank et al. 1999]. For training and evaluation they had 30 people, working in pairs, index 20 documents. Figure 7 shows key topics for one document and illustrates the inherent subjectivity of the task—the indexers achieved only 30% agreement with each other. Medelyan et al.'s automatic system, whose choices are shown as filled circles in the figure, obtained the same level of agreement and requires little training.

Although it has not been evaluated as such, Gabrilovich and Markovitch's [2007] ESA (Section 3.1) essentially performs topic indexing. For each document or text fragment it generates a weighted list of relevant Wikipedia concepts, the strongest of which should be suitable topic labels. Another approach that has not been compared to manually indexed documents is Schönhofen [2006], who uses Wikipedia categories as the vocabulary from which key topics are selected. Documents are scanned to identify the article titles and redirects they mention, and are represented by the categories that contain these articles—weighted by how often the document mentions the category title, its child article titles, and the individual words in them. Schönhofen did not compare the resulting categories with index topics, but instead used them to perform document categorization. Roughly the same results are achieved whether documents are represented by these categories or by their content in the standard way; combining the two yields a significant improvement.

Like document categorization, research in topic indexing builds solidly on related work but has been augmented to make productive use of Wikipedia. Significant gains have been achieved over the previous state of the art, although the results have not yet been evaluated as rigorously as in categorization. Medelyan et al. [2008] have directly compared their results against manually defined ground truth, but used a small dataset. To advance further, larger datasets need to be developed for evaluation and training.

## 5. INFORMATION EXTRACTION AND ONTOLOGY BUILDING

Whereas information retrieval aims to answer specific questions, information extraction seeks to deduce meaningful structures from unstructured data such as natural language text—though in practice the dividing line between the fields is not sharp. These structures are usually represented as relations. For example, from:

> *Apple Inc.'s world corporate headquarters are located in the middle of Silicon Valley, at 1 Infinite Loop, Cupertino, California.*

a relation *hasHeadquarters*(*Apple Inc., 1 Infinite Loop-Cupertino-California*) might be extracted. The challenge is to extract this relation from sentences that express the same information about *Apple Inc.* regardless of their actual wording. Moreover, given a similar sentence about other companies, the same relation should be determined with different arguments, e.g., *hasHeadquarters*(*Google Inc., Google Campus-Mountain View-California*). The next step after determining such relations would be to automatically organize them into a connected scheme, building a single machine-readable knowledge structure. Such organization attempts vary from producing simple thesauri through more complex taxonomies to comprehensive ontologies.

Section 5.1 begins with traditional information extraction approaches that apply methods developed before Wikipedia was recognized as something more than just a corpus: for them, any text represents a source of relations. The extraction process benefits from the encyclopedic nature of Wikipedia and its uniform writing style. Section 5.2 treats the determination of named entities and their type as a task of its own. Such work extracts information such as *isA*(*Portugal, Location*) and *isA*(*Bob Marley, Person*).

Section 5.3 turns to more ambitious approaches that see in Wikipedia's semi-structured parts and internal hyperlink structure (Sections 2.2.3, 2.2.4 and 2.2.5) the skeleton of a unified knowledge scheme. Ultimately, these researchers aspire to build a machine-readable ontology such as Cyc[22] [Lenat and Guha, 1990, Lenat, 1995] that captures the meaning of natural language as a whole using explicitly coded facts and rules, and/or a principled taxonomy that enables knowledge inheritance.

[22] http://www.opencyc.org

There is dispute concerning just what characterizes an ontology. What distinguishes it from (on the IT side) a mere database or (on the linguistic side) a mere controlled vocabulary? Some authors [e.g. Wilks, 2006] divide ontology research into two distinct areas: i) the field of 'knowledge representation', which descends from classical AI and aspires to make every aspect of the knowledge represented accessible to further inference, often resulting in major challenges with respect to inferential tractability; and ii) 'ontology engineering' understood merely as the development of shared conceptual schemes, for instance within a particular professional knowledge domain (biomedical ontologies[23] are a good example). Nevertheless, most agree that a formal ontology is a codification of the meanings of a set of concepts that is machine-readable to at least some extent. Building such a resource manually involves naming the concepts, representing and categorizing links between them, and (often) encoding key facts about them. Thus an ontologization of the concept *tree* should i) name it as a first-class object (to which equivalent terms such as the French *arbre* may be attached), ii) link it to closely-related concepts such as *leaf*, preferably with some indication that a leaf is part of a tree, and iii) be capable of representing facts like "There are no trees in the Antarctic."

Finally, Section 5.4 discusses attempts to use Wikipedia by Semantic Web researchers, whose goal is to transform knowledge sharing across the Internet.

## 5.1 Deriving relations from article text

Extracting semantic relations from raw text takes known relations as seeds, identifies patterns in their text—*X's * headquarters are located in * at Y* in the above example—and applies them to a large corpus to identify new relations. Phrase chunkers or named entity recognizers are applied to identify entities that appear in a sentence; intervening patterns are compared to the seed patterns; and when they match, new semantic relations are discovered. Culotta et al. [2006] summarize difficulties in this process:

- Enumeration over all pairs of entities yields a low density of correct relations, even when restricted to a single sentence.
- Errors in the entity recognition stage create inaccuracies in relation classification.

Wikipedia's structure helps combat these difficulties. Each article represents a particular concept that serves as a clearly recognizable *principal entity* for relation extraction from that article. Its description contains links to other, secondary, entities. All that remains is to determine the semantic relation between these entities. For example, the description of

[23] See for instance Open Biomedical Ontologies at http://www.obofoundry.org/

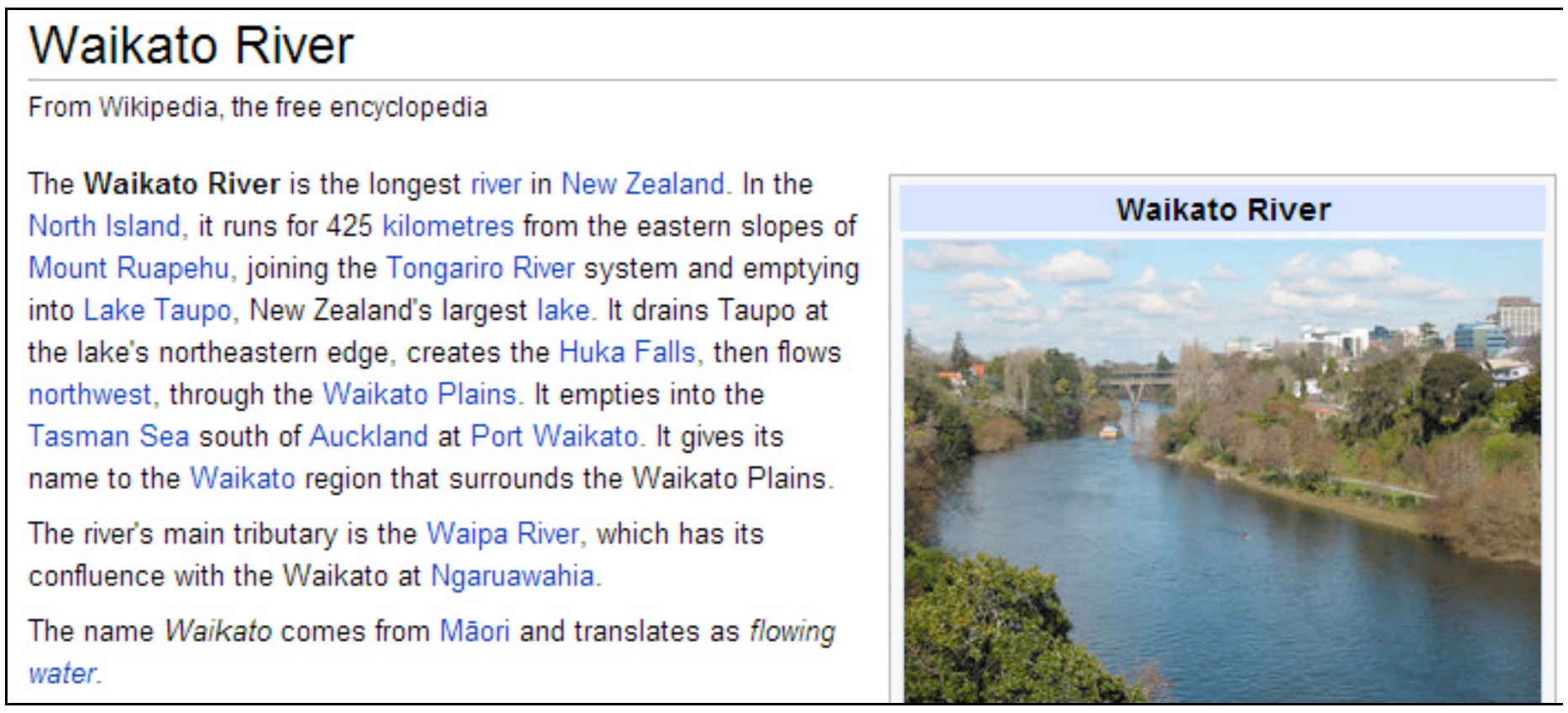
Waikato River

From Wikipedia, the free encyclopedia

The **Waikato River** is the longest river in New Zealand. In the North Island, it runs for 425 kilometres from the eastern slopes of Mount Ruapehu, joining the Tongariro River system and emptying into Lake Taupo, New Zealand's largest lake. It drains Taupo at the lake's northeastern edge, creates the Huka Falls, then flows northwest, through the Waikato Plains. It empties into the Tasman Sea south of Auckland at Port Waikato. It gives its name to the Waikato region that surrounds the Waikato Plains.

The river's main tributary is the Waipa River, which has its confluence with the Waikato at Ngaruawahia.

The name *Waikato* comes from Māori and translates as *flowing water*.

**Waikato River**

Figure 8. Wikipedia's description of the *Waikato River*.

the *Waikato River*, shown in Figure 8, links to entities like *river*, *New Zealand, Lake Taupo* and many others. Appropriate syntactic and lexical patterns can extract a host of semantic relations between these items.

Ruiz-Casado et al. [2005b] use WordNet for mining such patterns. Given two co-occurring semantically related WordNet nouns, the text that appears between them in Wikipedia articles is used to find relations missing from WordNet. But first the text is generalized. If the edit distance falls below a predefined threshold—i.e., the strings nearly match—those parts that do not match are replaced by a wildcard (*). For example, a generalized pattern *X directed the * famous|well-known film Y* is obtained from *X directed the famous film Y* and *X directed the well-known film Y*. Using this technique Ruiz-Casado et al. identify 1200 new relations with a precision of 61–69% depending on the relation type. In later work, Ruiz-Casado et al. [2006] extend this experiment by restricting Wikipedia pages to particular categories (*prime ministers*, *authors*, *actors*, *football players*, and *capitals*) before applying the patterns. Results vary wildly when the pages are combined into a single corpus, from 8% precision on the *player-team* relation to 90% for *death-year*, because of heterogeneity in style and mark-up of articles. But restricting to relevant categories yields a marked improvement, increasing to 93% when *player-team* patterns are applied solely to articles about football players.

Later approaches combine pattern extraction with syntactic parsing to improve coverage, because the same semantic relation can be expressed in different ways, i.e. *Chopin was great among the composers of his time* and *Chopin is a 19$^{th}$ Century composer.* Such patterns call for syntactic rather than lexical generalization. Herbelot and Copestake [2006] use a dependency parser to identify subject, object and their relationship in a sentence, regardless of word order. Their parser re-organizes a sentence into a series of minimal semantic trees whose roots correspond to lemmas in the sentence (see Figure 9). The same tree may be obtained for similar sentences. The results, which were restricted to articles describing animal species, are evaluated manually on a subset of 100 articles and automatically using a thesaurus. With three manually defined patterns recall was low: 14% at precision 92%. When patterns are extracted automatically recall improves to 37%; however precision drops to 65%.

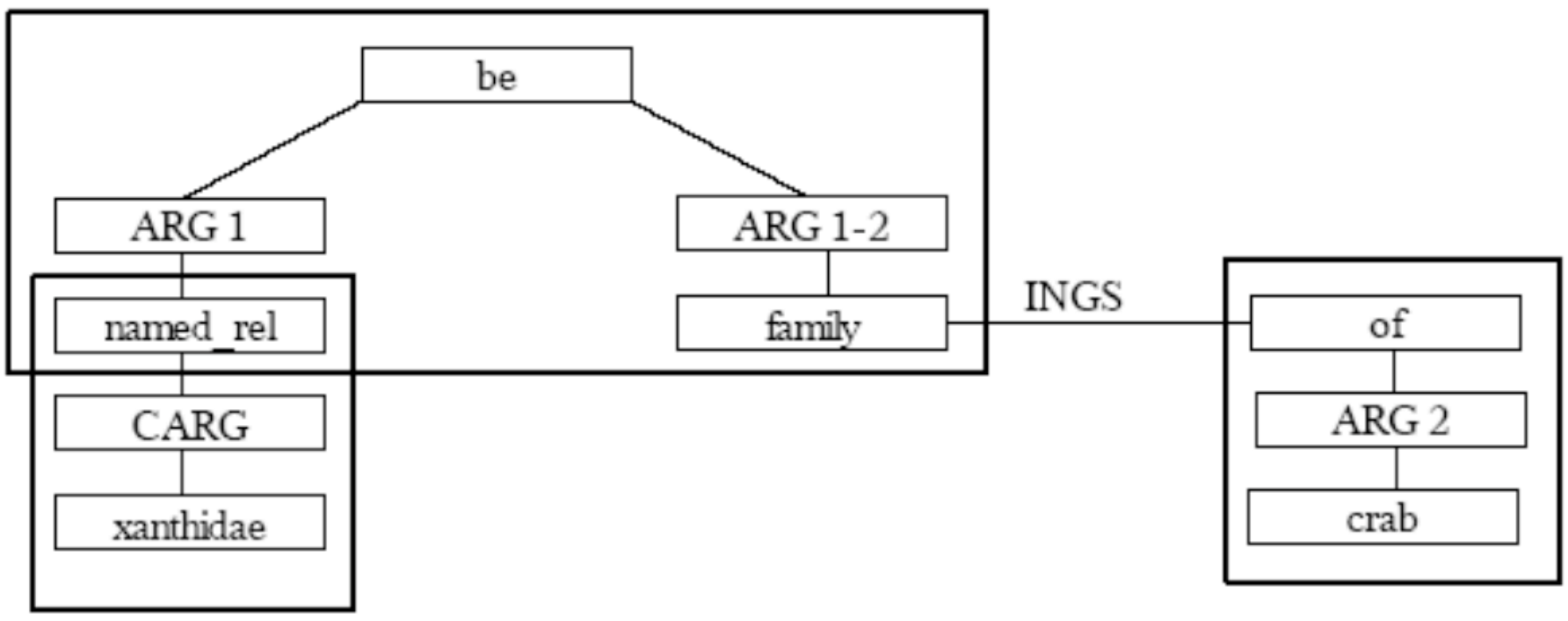

Figure 9. Output of the Robust Minimal Recursion Semantics analyzer for the sentence *Xanthidae is one of the families of crabs* [Herbelot and Copestake, 2006].

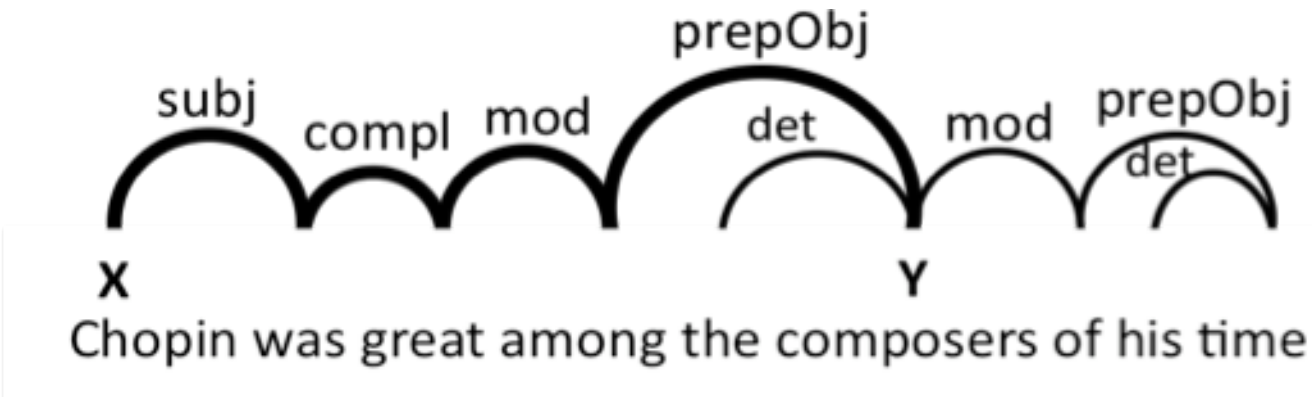

Figure 10. Example bridge pattern used in Suchanek et al. [2006].

Suchanek et al. [2006] use a context-free grammar for parsing. A pattern is defined by a set of syntactic links between two given concepts, called a *bridge*. For example, the bridge in Figure 10 matches sentences like the one above about *Chopin,* where *Chopin=X* and *composers=Y*. Machine learning techniques are applied to determine and generalize patterns that describe relations of interest from manually supplied positive and negative examples. The approach is evaluated on article sets with different degrees of heterogeneity: articles about composers, geography, and random articles. As expected, the more heterogeneous the corpus the worse the results, with best results achieved on composers for the relations *birthDate* (F-measure 75%) and *instanceOf* (F-measure 79%). Unlike Herbelot and Copestake [2006], Suchanek et al. show that their approach outperforms other systems, including a shallow pattern matching resource TextToOnto and the more sophisticated scheme of Chimiano and Volker [2005].

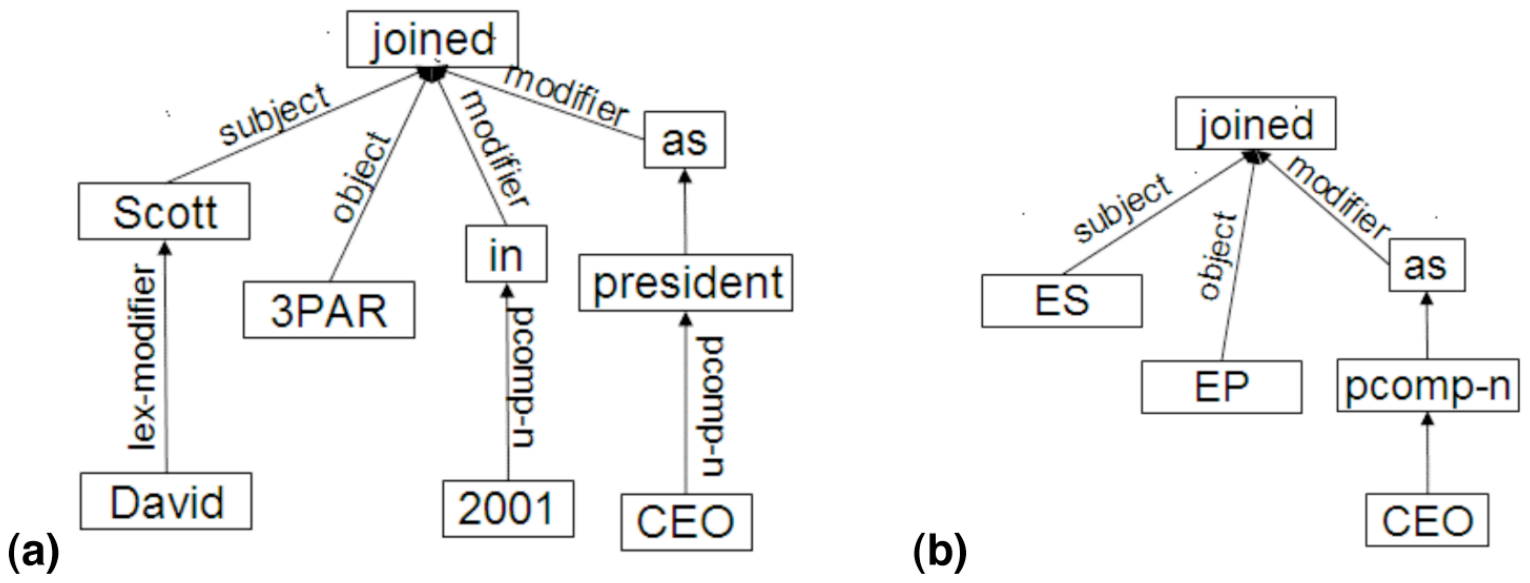

Figure 11. Example dependency parse in Nguyen et al. [2007a].

Nguyen et al. [2007a, 2007b], like Herbelot and Copestake [2006], use a dependency parser, but increase coverage using the OpenNLP toolkit's anaphora resolution.[24] For example, in an article about the software company 3PAR, phrases like *3PAR*, *manufacturer*, *it* and *company* are tagged as the same principal entity. All links are tagged as its secondary entities. Sentences with at least one principal and one secondary entity are analyzed by the parser. Like Suchanek et al. [2006], Nguyen et al. apply machine learning to generalize the trees, a task they call *subtree mining*. The dependency tree of Figure 11a is extracted from *David Scott joined 3PAR as CEO in January 2001* and is then generalized to match similar sentences (Figure 11b). The subtrees are extracted from a set of training sentences containing positive examples and applied as patterns to find new semantic relations. The scheme was evaluated using 3,300 manually annotated entities, 200 of which were reserved for testing. 6,000 Wikipedia articles, including 45 test articles, were used as the corpus. The approach achieved an F-measure of 38%, with precision much higher than recall, significantly outperforming two simple baselines.

Wang et al. [2007b] use selectional constraints in order to increase the precision of regular expressions without reducing coverage. They also extract positive seeds from infoboxes automatically. For example, the infobox field *Directed by* describes relation *hasDirector*(*FILM, DIRECTOR*) with positive examples <*Titanic, James Cameron*> and <*King Kong* (*2005*)*, Peter Jackson*>. They collect patterns that intervene between these entities in Wikipedia's text and generalize them into regular expressions like:

*X* (*is*|*was*) (*a*|*an*) * (*film*|*movie*) *directed by Y*.

Selectional constraints restrict the types of subject and object that can co-occur within such patterns. For example, Y in the pattern above must be a *director*—or at least a *person*. The labels specifying the types of entities implemented as features are derived from words that commonly occur in articles describing these entities. For example, instances of *ARTIST* extracted from a relation *hasArtist*(*ALBUM, ARTIST*) often co-occur with terms like *singer*, *musician*, *guitarist*, *rapper*, etc. To ensure better coverage, Wang et al. cluster such terms hierarchically. The relations *hasDirector* and *hasArtist* are evaluated independently on a sample of 100 relations extracted automatically from the entire Wikipedia, and were manually assessed by three human subjects. An unsupervised learning algorithm was applied, and the features were tested individually and together. The authors report precision and accuracy values close to 100%.

[24] *http://opennlp.sourceforge.net/*

The same authors investigate a different technique that does not rely on patterns at all [Wang et al. 2007c]. Instead, features are extracted from two articles before determining their relation:

- The first noun phrase and its lexical head that follows the verb *to be* in the article's first sentence (e.g., *comedy film* and *film* in *Annie Hall is a romantic comedy film*).
- Noun phrases appearing in corresponding category titles and the lexical heads.
- Infobox predicates, e.g. *Directed by* and *Produced by* in *Annie Hall*.
- Intervening text between two terms in sentences that use them both as a link.

For each pair of articles the distribution of these feature values is compared with that of positive examples. Unlike Wang et al. [2007b], no negative instances are used. A special learning algorithm (B-POL) designed for situations where only positive examples are available is applied. It identifies negative examples in unlabeled data automatically using a combination of a quick classifier and a more rigorous one. Four relations (*hasArtist*, *hasDirector*, *isLocatedIn* and *isMemberOf*) between 1,000 named entity pairs were evaluated by three human subjects. Best results were an F-measure of 80% on the *hasArtist* relation, which had the largest training set; the worse was 50% on *isMemberOf*.

Wu and Weld [2007, 2008] extract relations in order to improve Wikipedia infoboxes. Like Wang et al. [2007b, 2007c] they use their content as training data. Their system first maps infobox attribute-value pairs to sentences in the corresponding Wikipedia article using simple heuristics. Next, for each attribute it creates a sentence classifier that uses the sentence's tokens and their part of speech tags as features. Given an unseen Wikipedia article, a document classifier analyzes its categories and assigns an infobox class, e.g. 'U.S. counties'. Next, a sentence classifier is applied to assign relevant infobox attributes. Extracting values from sentences is treated as a sequential data-labeling problem, to which Conditional Random Fields are applied. Precision and recall are measured by comparing generated infoboxes against existing ones. The authors manually judged the attributes produced by their system and by Wikipedia authors. Precision ranged from 74 to 97% at recall levels of 60 to 96% respectively, depending on the infobox class. Precision was around 95% on average and more stable across the classes; recall was significantly better on most classes but far worse on others. This paper will be discussed further in Section 5.4 on the Semantic Web.

In later work Wu et al. [2008] address problems in their approach in the following way. To generate complete infobox schemata for articles of rare classes, they refer to WordNet's ontology and aggregate attributes from parents to their children classes. For example, knowing that *isA*(*Performer, Person*), the infobox for *Performers* receives the formerly missing field *BirthPlace*. To provide additional positive examples, they seek new sentences describing the same attribute-values pairs by applying TextRunner [Banko et al. 2007] to the web. Given a new entity for which an infobox is to be generated, Google search is used to retrieve additional sentences describing it. The combination of these techniques improves recall by 2–9 percentage points while maintaining or increasing precision. These results are the most complete and impressive so far.

Most of these approaches capitalize on Wikipedia's encyclopedic nature, using it as a corpus for extracting semantic relations. Simple pattern matching is improved by the use of parsing [Suchanek et al. 2006], anaphor resolution [Nguyen et al. 2007a, 2007b], selectional constraints [Wang et al. 2007b] and lexical analysis [Wu and Weld 2007, 2008]; however, the exact contribution of each is unclear. Many of the methods seem complementary and could be combined into a single approach, but experiments have not yet been reported. There is a clear shift from using patterns to the use of features in machine learning approaches [Nguyen et al. 2007b, Wang et al. 2007c, Wu and Weld. 2007, 2008]. Wikipedia itself is used as a source of training examples [Wang et al. 2007b, Wu and Weld 2007], instead of defining them manually. Wu et al. [2008] demonstrate that performance can be boosted by retrieving additional content from the web.

It would be helpful to compare the approaches on the same data set. Of course, the researchers would have to reach a consensus on what relations to extract, and at this point there are merely arbitrary overlaps in some relations (*isMemberOf*, *instanceOf*, *hasDirector*). There is little cross-pollination between this research and that in Section 5.3, where semantic relations are extracted directly from Wikipedia's structure—like category links and infoboxes. Section 5.3 will show that these contain a wealth of semantic relations, outnumbering those appearing in the article text. On the positive side, unlike those in Section 5.3 the techniques surveyed in this section generalize to the entire web. Article text is the primary source of relations, and infoboxes are used to enhance the extraction of meaningful nuggets. It is easy to imagine bootstrapping these results to the rest of the web, which we return to in Section 5.4.

### 5.2 Typing named entities

Infoboxes for entities of the same kind share similar characteristics—for example, *Apple Inc*, *Microsoft* and *Google* share the fields *Founded*, *Headquarters*, *Key People* and *Products*—but Wikipedia does not state that they belong to the same type of named entity, in this case *company*. This would greatly help tasks such as information retrieval and question answering (Section 4). This section covers research that classifies articles into predefined classes representing entity types. The results are semantic relations of a particular kind, e.g. *isA*(*London, Location*), whose subject is a named entity and object the particular type of entity.

Toral and Muñoz [2006] extract named entities from the Simple Wikipedia using WordNet's noun hierarchy. Given an entry—*Portugal*—they extract the first sentence of its definition—*Portugal is a country in the south-west of Europe*—and tag each word with its part of speech. They assign nouns their first (i.e., most common) sense from WordNet and move up the hierarchy to determine its class, e.g., *country* → *location*. The most common class appearing in the sentence determines the class of the article (i.e., entity). The authors achieve 78% F-measure on 400 *locations* and 68% on 240 *persons*. They do not use Wikipedia's structural features but mention this as future work.

Buscaldi and Rosso [2007b] address the same task but concentrate on geographical locations. Unlike Toral and Muñoz [2006], they analyze the entire text of each article. To determine whether it describes a location they compare its content with a set of keywords extracted from glosses of locations in WordNet using the Dice metric and cosine coefficient; they also use a multinomial Naïve Bayes classifier trained on the Wikipedia XML corpus [Denoyer and Gallinari 2006]. When evaluated on data provided by Overell and Rüger [2007] (Section 3.2.2) they find that cosine similarity outperforms both the WordNet-based Dice metric and Naïve Bayes, achieving an F-measure of 53% on full articles and 65% on first sentences. However, the results fall short of Overell and Rüger's [2006], and the authors conclude that the content of articles describing locations is less discriminative than other features like geographical coordinates.

Section 3.2.2 discussed how Overell and Rüger [2006, 2007] analyze named entities representing geographic locations, mapping articles to place names listed in a gazetteer. It also described another group of approaches that recognize named entities in raw text and map them to articles. Apart from these, little research has been done on determining the semantic types of named entities. It is surprising that both techniques described above use WordNet as a reference for the entities' semantic classes instead of referring to Wikipedia's categories. For example, three companies mentioned above belong to subcategories of *Category:Companies* and *Portugal* is listed under *Category:Countries*.

Some of the approaches described in the following section use exactly this information to mine *isA* relations, but are not restricted to named entities. Moreover, neither of the techniques discussed here utilizes the shared infobox fields. Annotating Wikipedia with entity labels looks like low-hanging fruit: we expect more advances in the future.

### 5.3 Ontology building

We have described how to extract facts and relations from Wikipedia individually. The present section goes farther, using Wikipedia as the source of a single comprehensive knowledge base, an ontology that gives an alternative to manually created resources such as Cyc and WordNet. The three main approaches we see so far are i) extracting and/or labeling Wikipedia's category relations, ii) extracting and organizing its infobox assertions, iii) adding information mined from Wikipedia to existing manually-built resources. We discuss these approaches and compare their coverage and depth.

Chernov et al. [2006] were among the first to investigate whether links between Wikipedia categories bear semantic meaning. They found that the hyperlink connectivity between the articles in two categories correlates with the semantic relatedness of those categories. Nakayama et al. [2007a, 2007b, 2008] extend this idea by building a large general-purpose thesaurus solely from the hyperlink structure. The result contains 1.3M concepts with a semantic relatedness weight assigned to every concept pair. They suggest that the thesaurus may easily be upgraded to a full-blown ontology by 'typing' the generic relatedness measures between concepts into more traditional ontological relations such as *isA* and *part of*, though details are sketchy.

An ontology project closer to classical knowledge representation is YAGO, *Yet Another Great Ontology* [Suchanek et al. 2007]. This creates a giant taxonomy by mapping Wikipedia's leaf categories onto the WordNet taxonomy of synsets and adding the articles belonging to those categories as new elements. Each category's lexical head is extracted—*people* in *Category:American people in Japan* —then sought in WordNet. If there is a match, it is chosen as the class for this category. This scheme extracts 143,000 *subClass* relations—in this case, *subClass*(*American people in Japan, person/human*). If more than one match is possible, word sense disambiguation is required (Section 3.2). The authors experimented with mapping a category's subcategories to WordNet and choosing the sense that corresponds to the smallest resulting taxonomic graph. However, they claim that this technique does not perform as well as choosing the most frequent WordNet synset for a given term (the frequency values are provided by WordNet), an observation that seems inconsistent with findings by other authors [e.g. Medelyan and Milne 2008] that the most frequent sense is not necessarily the intended one (Section 3.2.3). YAGO's use of the manually-created WordNet taxonomy neatly bypasses the poor ontological quality of Wikipedia's category structure. It also avoids Ruiz-Casado et al.'s [2005b] problem of omitting Wikipedia concepts whose titles do not appear in WordNet, although it still misses proper names with WordNet *synonyms*—e.g. the programming language *Python* and the movie *The Birds.*

| Relation | Domain | Range | Number of facts |
| --- | --- | --- | --- |
| subClassOf | class | class | 143,210 |
| type (isA) | entity | class | 1,901,130 |
| context | entity | entity | 40,000,000 |
| describes | word | entity | 986,628 |
| bornInYear | person | year | 188,128 |
| diedInYear | person | year | 92,607 |
| establishedIn | entity | year | 13,619 |
| locatedIn | object | region | 59,716 |
| writtenInYear | book | year | 9,670 |
| politicianOf | organization | person | 3,599 |
| hasWonPrize | person | prize | 1,016 |
| means | word | entity | 1,598,684 |
| familyNameOf | word | person | 223,194 |
| givenNameOf | word | person | 217,132 |

Table 5. Size of YAGO (facts).

The authors define a mixed suite of heuristics for extracting further relations to augment the taxonomy. For instance, a name parser is applied to all personal names to identify given and family names, adding 440,000 relations like *familyNameOf(Albert Einstein, "Einstein")*. Many heuristics make use of Wikipedia category names, extracting relations such as *bornInYear* from subcategories of categories ending with *birth* (e.g., *1879 birth*), or *locatedIn* from a category like *Cities in Germany*. This yields 370,000 non-hierarchical, non-synonymous relations. Manual evaluation of sample facts shows 91–99% accuracy, depending on the relation.

From an ontology-building perspective, parsing category names is a real step forward, though only a tiny subset have so far been resolved. For instance, YAGO does not recognize widespread patterns such as "X by Y" (e.g., *Persons by continent*, *Persons by company*) as Ponzetto and Strube [2007b] do (below). Also added are 2M synonymy relations generated from redirects, 40M context relations generated from cross-links between articles, and 2M type relations between categories considered as classes and their articles considered as entities (though this constitutes a questionable ontological short-cut—*Article:Cat* and *Category:United Nations* are just two negative examples). Overall YAGO claims to know 20M facts about 2M entities (Table 5). [25]

[25] YAGO can be queried online: *http://www.mpi-inf.mpg.de/~suchanek/downloads/yago/*

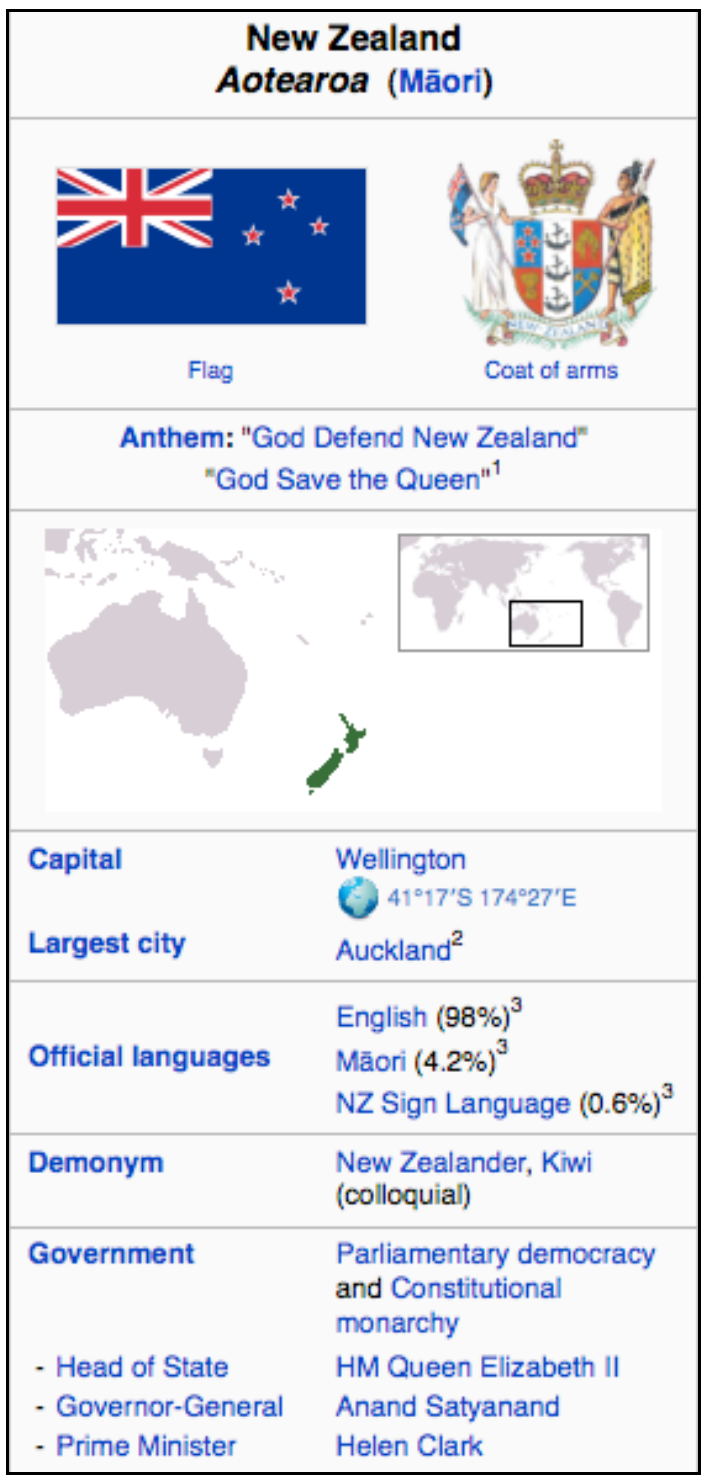

```
{{ Infobox Country or territory |

native_name = New Zealand |
…
capital = [[Wellington]] |

latd = 41 | latm = 17 | latNS = S |
longd = 174 | longm = 27 | longEW = E |

largest_city = [[Auckland]] |

official_languages =
        [[New Zealand English|English]] (98%)
        [[Māori language|Māori]] (4.2%)
        [[New Zealand Sign Language|NZ Sign
        Language]] (0.6%) |

demonym = [[New Zealand People|New
        Zealander]],[[Kiwi (people)|Kiwi]] |

government_type =
        [[Parliamentary democracy]] and
        [[Constitutional monarchy]]
…}}
```

Figure 12. Wikipedia infobox on New Zealand.

Another notable feature of YAGO considered from a knowledge representation perspective is that it includes a logic-based representation language and a basic data model of entities and binary relations, with a small extension to represent relations between facts (such as transitivity). This gives it formal rigor—the authors even provide a model-theoretic semantics—and the expressive power of a rich version of Description Logic. In terms of inferential tractability it compares favorably with the hand-crafted (higher-order logic) Cyc. An online SPARQL interface allows logically complex queries. For instance, when asked for *billionaires born in the USA* it produced two (though it missed *Bill Gates*—the system does not cover Wikipedia's structured data fully). The project will be integrated with the latest version of OWL. The authors claim to have already noticed a positive feedback loop whereby as more facts are added, word senses can be disambiguated more effectively in order to correctly identify and enter further facts. This was a long-standing ambition of researchers in knowledge representation [Lenat 1995], though claims of its achievement often turned out to be premature.

An even larger scale, but less formally structured, relation extraction project is DBpedia [Auer et al. 2007; Auer and Lehmann 2007]. This transforms Wikipedia's structured and semi-structured information (most notably infoboxes) into a vast set of RDF triples. Figure 12 shows the infobox from the *New Zealand* article; on the right is the

| Dataset | Description | Triples |
|---|---|---|
| Page links | Internal links between DBpedia instances derived from the internal pagelinks between Wikipedia articles | 62 M |
| Infoboxes | Data attributes for concepts that have been extracted from Wikipedia infoboxes | 15.5 M |
| Articles | Descriptions of all 1.95M concepts within the English Wikipedia. Includes titles, short abstracts, thumbnails and links to the corresponding articles | 7.6 M |
| Languages | Additional titles, short abstracts and Wikipedia article links in 13 other languages. | 5.7 M |
| Article categories | Links from concepts to categories using SKOS | 5.2 M |
| Extended abstracts | Additional, extended English abstracts | 2.1 M |
| Language abstracts | Extended abstracts in 13 languages | 1.9 M |
| Type information | Inferred from category structure and redirects by the YAGO (“yet another great ontology”) project [Suchanek et al. 2007] | 1.9 M |
| External links | Links to external web pages about a concept | 1.6 M |
| Categories | Information which concept is a category and how categories are related | 1 M |
| Persons | Information about 80,000 persons (date and place of birth etc.) represented using the FOAF vocabulary | 0.5 M |
| External links | Links between DBpedia and Geonames, US Census, Musicbrainz, Project Gutenberg, the DBLP bibliography and the RDF Book Mashup | 180 K |

Table 6. Content of DBpedia [Auer et al. 2007].

Wiki mark-up used to create it. Extracting information from infoboxes is not trivial. There are many different templates, with a great deal of redundancy—for example, *Infobox_film*, *Infobox Film*, and *Infobox film*. Recursive regular expressions are used to parse relational triples from all commonly used Wikipedia templates containing several predicates. The templates are taken at face value; no heuristics are applied to verify their accuracy. The Wikipedia URL of each entity is recorded as a unique identifier.

Unlike YAGO there is no attempt to place facts in the framework of an overall taxonomic structure of concepts. Links between categories are merely extracted and labeled with the relation *isRelatedTo*. As with YAGO, Wikipedia categories are treated as classes and articles as individuals. In DBPedia’s case this is particularly problematic given that many articles have corresponding categories (e.g. *New Zealand*), so presumably the two receive entirely different identifiers and their semantic relationship is obscured. (Auer and Lehmann do not say what happens in this case.) The resulting DBpedia dataset contains 115,000 classes and 650,000 individuals sharing 8,000 different types of semantic relations. A total of 103M triples are extracted, far surpassing any other scheme in size.[26] Like YAGO, the dataset can be queried via SPARQL and Linked Data, and connects with other open datasets on the web. Table 6 summarizes its content.

[26] Further information, and the extracted data, can be downloaded from *http://www.dbpedia.org*

The unsurpassed quantity of information in DBpedia is a wonderful resource for the research community, particularly given its multilingual character, and it is becoming something of a hub for free large-scale data-repositories. (This will be discussed further in section 5.4 on the Semantic Web.) However, it's worth noting that as an *ontology* DBpedia falls short on some traditional expectations. First, since as mentioned there is little or no connection between its facts, such as would be provided by an inheritance hierarchy of concepts and/or a formally defined ontology language, it would seem that many semantic relations amongst its triples will go unrecognized.

Second, although no formal evaluation of quality is provided, a quick manual inspection reveals that large sections of the data have limited ontological value taken as-is. For instance, 60% of the RDF triples are relatively trivial semantic relations derived from Wikipedia's link structure (e.g. *hasCategory*, *Book*, *Category:Documents*); only 15% are taken directly from infoboxes. Amongst those are the many obviously redundant relations. Finally, some individual infobox-derived relations contain very poor quality data, presumably caused by erroneous parsing of inconsistent values—for instance, *keyPeople* assertions contain values such as "CEO", which is a role rather than a person, or "Bob", which is underspecified. Unlike other approaches, DBpedia relies entirely on the accuracy of Wikipedia's contributors, and Auer and Lehmann suggest guidelines for authors to improve the quality of infoboxes with time.

Lately (November 2008) the project has attempted to remedy the lack of formal structure by releasing the DBpedia Ontology,[27] which has been "manually created based on the most commonly used infoboxes within Wikipedia." It currently includes around 170 classes, which form a subsumption hierarchy and have 940 properties and about 882,000 instances—much smaller than YAGO's claimed 2M. Its creators have addressed the redundancy in the raw DBpedia data—thus 350 Wikipedia templates have been reduced to 170 ontology classes and 2350 template relations have been mapped to just 940 ontology relations. They also endeavour to produce clearly defined datatypes for all property values. Again this is valuable new resource, although quick manual inspection reveals some inaccuracies. For instance, template names seem to be used to generate *isa* assertions on the entities described by articles that contain the template, which leads to errors such as asteroids being categorized as *Planet*. No formal evaluation of this resource has so far been reported.

Work at the European Media Lab Research Institute (EMLR) takes another approach to building an ontology from Wikipedia, where the basic building block is Wikipedia's category links. Ponzetto and Strube [2007] begin by identifying and isolating *isA* relations from the rest of the category links (which they call *notIsA*). Here *isA* is thought of as subsuming relations between two classes—*isSubclassOf*(*Apples, Fruit*)—and between an instance and its class—*isInstanceOf*(*New Zealand, Country*). Several steps are applied. One of the most accurate matches the lexical head and modifier of two category names. Sharing the same head indicates *isA*, e.g., *isA*(*British computer scientist, Computer scientist*). Modifier matching indicates *notIsA*, e.g., *notIsA*(*Islamic mysticism, Islam*). Another method uses co-occurrence statistics of categories within patterns to indicate hierarchical and non-hierarchical relations. For noun phrases A and B, *A such as B* (e.g., *fruit such as apples*) indicates *isA*, and the intervening text can be generalized to *like*, "*, especially*", and so on. Similarly, *A are used in B* (*fruit are used in cooking*) indicates *notIsA*. This technique induces 100,000 *isA* relations from Wikipedia. Comparison with relations manually assigned to concepts with the same lexical heads in ResearchCyc shows that the labeling is highly accurate, depending on the method used, and yields an overall F-measure of 88%. Ponzetto and Strube [2007a] apply the induced taxonomy to natural language processing tasks such as co-reference resolution.

Zirn et al. [2008] divide the derived *isA* relations into *isSubclassOf* and *isInstanceOf*, the two types mentioned in the previous paragraph. Instead of YAGO and DBpedia's assumption that all categories are classes and all articles are instances, the EMLR group seeks automated methods to determine their status on a case-by-case basis. Two of their methods assume that all named entities are instances and thus related to their categories by *isInstanceOf*. One uses a named entity recognizer, the other a heuristic based on capitalization in the category title (though this only works for multi-word titles). Pluralization is also considered. Further methods include heuristics such as: If a category has at least one hyponym that has at least two hyponyms, it is a class. Evaluation against 8,000 categories listed in ResearchCyc as individuals (instances) and collections (classes) shows that the capitalization method is best, achieving 83% accuracy; however, combining all methods into a single voting scheme improves this to 86%. The taxonomy derived from this work is available in RDF Schema format.[28]

Nastase and Strube [2008] begin to address the *notIsa* domain, extracting non-taxonomic relations from Wikipedia by parsing category titles. It is worth noting that they are no longer merely labeling links in the category network but deriving entirely new relations between categories, articles and terms extracted from category titles. Explicit unitary relations are extracted—for example, from the category title *Queen* (*band*) *members* they infer the *memberOf* relation from articles in that category to the article for the band, e.g. *memberOf*(*Brian May, Queen* (*band*)). Explicit binary relations are also extracted—for example, if a category title matches the pattern *X* [*VBN IN*] *Y* (e.g. *Movies directed by Woody Allen*), the verb phrase is used to 'type' a relation between all articles assigned to the category and the entity *Y* (*directedBy*(*Annie Hall*, *Woody Allen*)), while the class *X* is used to further type the articles in the category (*isA*(*Annie Hall, Movie*)).

[27] http://wiki.dbpedia.org/Ontology
[28] *http://www.eml-r.org/english/research/nlp/download/wikitaxonomy.php*

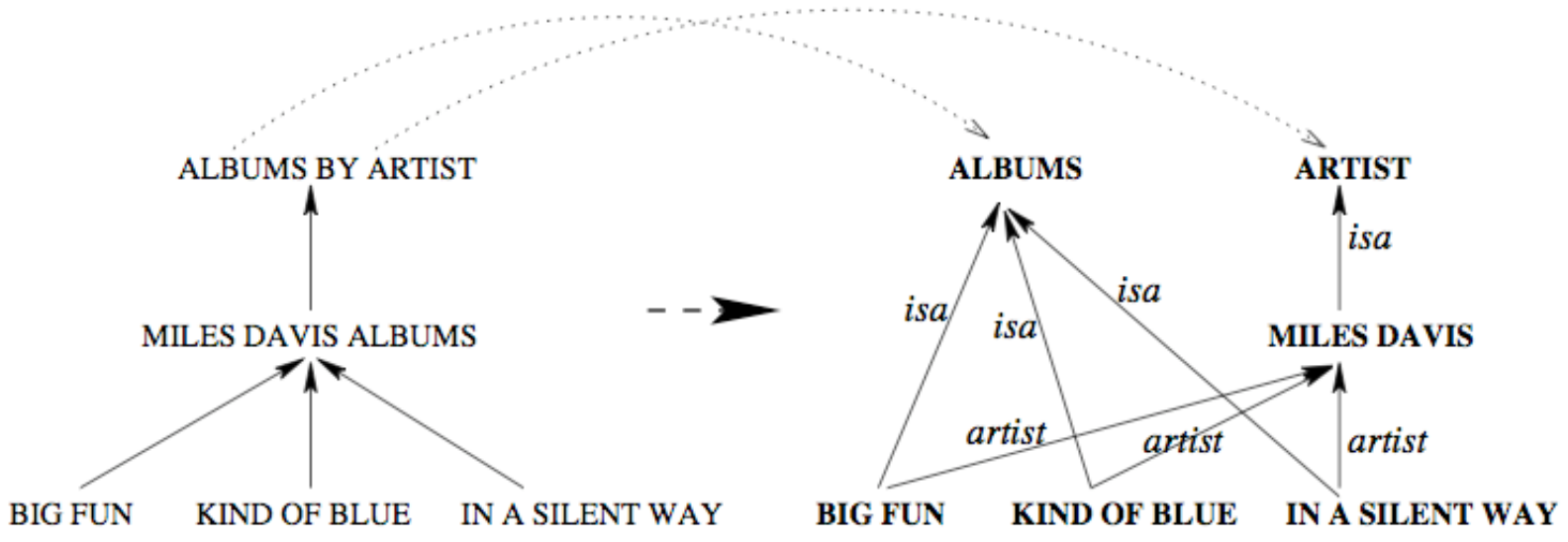

Figure 13. Relations inferred from BY categories [Nastase and Strube 2008].

Particularly sophisticated is their derivation of entirely implicit relations from the common *X by Y* pattern in Wikipedia category names, which facets much of the category structure (e.g. *Writers By Nationality, Writers by Genre, Writers by Language*). Given the category title *Albums By Artist,* they not only label all articles in the category *isA*(*X, Album*)*,* but also find subcategories pertaining to particular artists (e.g. *MilesDavis, Albums*), locate the article corresponding to the artist, label the entity as an artist, (*isA*(*MilesDavis, Artist*)) and label all members of the subcategory accordingly (*artist*(*KindOfBlue MilesDavis*)) as illustrated in Figure 13. They identify a total of 3.4M *isA* and 3.2M *spatial* relations, along with 43,000 *memberOf* relations and 44,000 other relations such as *causedBy* and *writtenBy*. Evaluation with ResearchCyc was not meaningful because of little overlap in extracted concepts—particularly named entities. Instead, human annotators analyzed four samples of 250 relations; precision ranged from 84–98% depending on relation type. Overall Nastase and Strube claim to add 9M new facts to their existing taxonomy of 105,000 categories, about twice the size of YAGO. They promise to release a new ontology containing these facts soon.

Another resource worth mentioning is Freebase,[29] the collaborative knowledge base produced by Metaweb. Like the previous structures, it contains many concepts and relations that have been automatically mined from Wikipedia. These are merged with other resources such as MusicBrainz (an extensive ontology of musicians, albums, and songs) and the Notable Names Database. Registered users can also contribute directly. Unlike the previous structures, it contains many images, and is loosely organized as a set of 4M 'topics' to which users add links associatively (thus creating a mixture of strict and informal relations). While the knowledge in Freebase is freely available (via browsable interfaces, database dumps, and APIs), it is intended to generate revenue through potential applications to advertising and search. We could find little information on the algorithms that generate it. Neither can we comment on its accuracy and coverage, because we know of no formal evaluation. Regardless, it is encouraging to see a large-scale knowledge base courting an active community of contributors. Freebase is used in a growing number of projects, notably the natural language search engine PowerSet.

[29] *http://www.freebase.com*

| | Ontology | Entities | Facts |
|---|---|---|---|
| Manually created | SUMO | 20,000 | 60,000 |
| | WordNet | 117,597 | 207,016 |
| | OpenCyc | 47,000 | 306,000 |
| | ResearchCyc | 250,000 | 2,200,000 |
| | DBpedia Ontology | 820,000 | Est. ~1M |
| Automatically derived | YAGO | 2M | 20M |
| | DBpedia | 2.6M | 103M |
| | EMLR[2008] | 1.7M | 11M |

Table 7. Relative size of ontologies (January 2009).

The resources presented in this section—YAGO, DBpedia, EMLR's taxonomy, and FreeBase—all have the same goal: to create an extensive, accurate, general knowledge base. The techniques used differ radically. YAGO combines Wikipedia's leaf categories (and their instances) with WordNet's taxonomy, embellishing this structure with further relations. DBpedia dumps Wikipedia's structured and semi-structured information with little further analysis. EMLR perform a sophisticated differentiation or 'typing' of category links, followed by an analysis of category titles and the articles contained by those categories to derive further relations.

As a result, the information extracted varies. Whereas Suchanek et al. [2007] extract the relation *writtenInYear*, Nastase and Strube [2008] detect *writtenBy* and Auer and Lehmann [2007] generate *written*, *writtenBy*, *writer*, *writers*, *writerName*, *coWriters*, as well as their case variants. Table 7 compares the sizes of sufficiently formally structured ontologies. Evaluation is still patchy and ad hoc, and quality is a major concern for the cruder automated ontology-building methods. For instance, distinguishing between instances and classes amongst Wikipedia concepts is difficult to solve automatically, but difficult to avoid given the widespread use of class hierarchies to structure ontologies. There has so far been little comparison of these approaches or attempts to integrate them, and no cross-pollination with the clearly complementary work described in Section 5.1.

## 5.4 Wikipedia and the Semantic Web

Here we consider research projects which seek to contribute to the broad-ranging and ambitious research project known as the Semantic Web, spearheaded by the World Wide Web Consortium (W3C). Its goal is to add metadata to Web documents to enable Web searches to access semantically enriched rather than unstructured material, thereby allowing computers to "become much better able to process and 'understand' the data that they merely display at present" [Berners-Lee et al. 2001]. (For an overview of the project's original goals, see also [Fensel et al. 2002].) A crucial plank of the project is getting the world involved in marking up web pages semantically, and this 'political' challenge has proved as much of an obstacle as the many technical challenges its developers have confronted [Legg 2007].

Wikipedia can be seen as the largest available semantically marked up corpus. For example: important phrases that appear in the definition of articles in Wikipedia are explicitly linked to other articles that describe their meaning; each article is assigned to one or more semantic categories and infobox templates encode typical attributes of concepts of the same kind. With such characteristics, Wikipedia can be seen not only as a mini-model of the Semantic Web, but also as a prototype of a tool designed to mark up at least some of the meaning in the entire web. Several methods have been proposed to evolve Wikipedia into a more structured and fully machine-readable repository of meaning.

Krötzsch et al. [2005, 2007] and Völkel et al. [2006] develop the idea of 'link typing' in greater detail than Nakayama et al. [2007a, 2007b, 2008] in Section 5.3. Rather than creating a new stand-alone resource, they plan to label Wikipedia's own hyperlink structure. Noting the profusion of links between articles, all indicating some form of semantic relatedness, they claim that categorizing them would be a simple, unintrusive way of rendering large parts of Wikipedia machine-readable. For instance, the link from *leaf* to *plant* would be labeled *partOf*, that from *leaf* to *organ kindOf*, and so on. As categorizing all hyperlinks would be a significant task, they recommend introducing a system of link types and encouraging Wikipedia editors to start using them, and to suggest further types. This raises interesting usability issues. Given that ontology is specialist knowledge (at least as traditionally practiced by ontological engineers), it might be argued that disaster could result from Wikipedia's uniquely democratic editing model. On the other hand, one might ask why this is any different to other specialist additions to Wikipedia (e.g. cell biology, Scottish jazz musicians), whose contributors show a remarkable ability to self-select, yielding surprising quality control. Perhaps the most tricky characteristic of ontology is that, unlike specialist topics such as cell biology, people think they are experts in it when in fact they are not. At any rate, as this research is essentially a proposal for Wikipedia's developers to add further functionality, its results cannot yet be evaluated.

Wu and Weld [2007, 2008] hope to kick-start the Semantic Web by marking up Wikipedia semantically using a combination of automated and human processes. They explore machine-learning techniques for completing infoboxes by extracting data from article text, constructing new infoboxes from templates where appropriate, rationalizing tags, merging replicated data using microformats, disambiguating links, adding additional links, and flagging items for verification, correction, or the addition of missing information. As with Krötzsch et al., it will be interesting to see whether Wikipedia editors will be eager to work on the collaborative side of this project, and also how effective they are.

Yet how much nearer does this bring us to Tim Berners Lee's original goal of allowing computers to *understand* the data that they merely display at present? It depends what is meant by 'understand'. If we only require that the terms within web documents are tagged with an appropriate Wikipedia article, then the *wikification* (discussed in Section 3.1) already qualifies. However many researchers have in mind a deeper understanding which would enable inferencing over marked-up data to deduce further consequences. This was certainly part of Tim Berners-Lee's original vision. From that perspective, an intimidating amount of work on extracting and analyzing Wikipedia's semantics remains. To be more precise, great progress towards the Semantic Web vision has been made with respect to named entities, for all that is needed to establish shared meaning for them is a shared URI. General concepts like *tree* are more tricky. Wikipedia certainly contains a wealth of semantic information regarding such concepts, but we have seen throughout this survey that there is little consensus on how to extract and analyze it, let alone inference over it.

As a semi-structured resource, Wikipedia sits somewhere between the chaotic and—as far as machines are concerned—utterly incomprehensible Web of today and the one envisioned by Tim Berners-Lee. In some sense, most of the work described in this survey aims to push this small subset of the web towards the imagined ideal, or to pragmatically make use of what we have already. Comparatively little has been done to generalize and bootstrap Wikipedia with the aim of understanding the rest of the web. In short, Wikipedia is long way from giving us the Semantic Web in miniature, let alone in full.

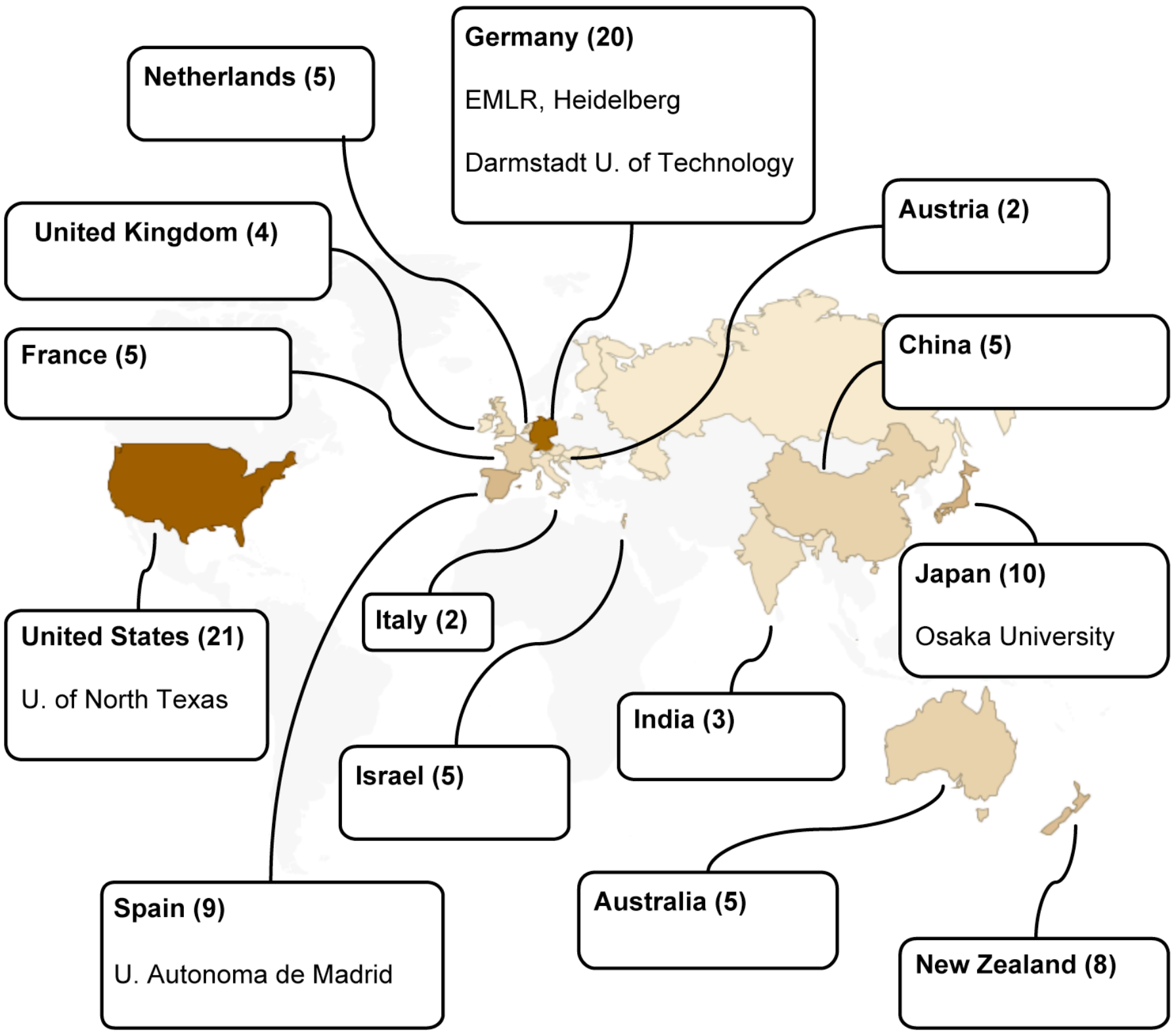

Figure 14. Countries and institutions with significant research on mining meaning from Wikipedia

## 6. PEOPLE, PLACES AND RESOURCES

Wikipedia began with the goal of distributing a free encyclopedia to every person on the planet, in their own language. It is reassuring to see the research community that benefits so much from Wikipedia maintaining the same international perspective. The research described in this survey is scattered across the globe. Figure 14 shows prominent countries and institutions at the time of writing (mid-2008, as stated at the outset).

US and Germany are the largest contributors. In the US, research is spread across many institutions. The University of North Texas, who work with entity recognition and disambiguation, produced the *wikify* system. In the Pacific Northwest, Microsoft Research focuses on named entity recognition, while the University of Washington extracts semantic relations from Wikipedia's infoboxes. German research is more localized geographically. EMLR works on relation extraction, semantic relatedness, and co-reference resolution; Darmstadt University of Technology on semantic relatedness and analyzing Wikipedia's structure. The Max Planck Institute produced the YAGO ontology; they collaborate with the University of Leipzig, who produced DBpedia. The University of Karlsruhe have focused on providing users with tools to add formal semantics to Wikipedia.

Spain is Europe's next largest contributor. Universidad Autonoma de Madrid extract semantic relations from Wikipedia; Universidad Politecnica de Valencia and Universidad de Alicente both use it to answer questions and recognize named entities. The Netherlands, France, and UK are each represented by a single institution. The University of Amsterdam focusses on question answering; INRIA works primarily on entity ranking, and Imperial College on recognizing and disambiguating geographical locations.

The Israel Institute of Technology have produced widely cited work on semantic relatedness, document representation and categorization. They developed the popular technique of Explicit Semantic Analysis.

Hewlett Packard's branch in Bangalore puts India on the map with document categorization research. In China, Shanghai's Jiatong University works on relation extraction and category recommendation. In Japan, the University of Osaka has produced several open source resources, including a thesaurus and a bilingual (Japanese–English) dictionary. The University of Tokyo, in conjunction with the National Institute of Advanced Industrial Science and Technology, has focused on relation extraction.

New Zealand and Australia are each represented by a single institution. Research at the University of Waikato covers entity recognition, query expansion, topic indexing, semantic relatedness and augmenting existing knowledge bases. RMIT in Melbourne have collaborated with INRIA's work on entity ranking.

Table 8 summarizes tools and resources, along with brief descriptions and URLs. This is split into tools for accessing and processing Wikipedia, demos of Wikipedia mining applications, and datasets that have been generated from Wikipedia.

| **Processing tools** | |
|---|---|
| JWPL Java Wikipedia Library | API for structural access of Wikipedia parts such as redirects, categories, articles and link structure. [Zesch et al. 2008] *http://www.ukp.tu-darmstadt.de/software/jwpl/* |
| WikiRelate! | API for computing semantic relatedness using Wikipedia [Strube and Ponzetto 2006; Ponzetto and Strube 2006] *http://www.eml-research.de/english/research/nlp/download/ wikipediasimilarity.php* |
| Wikipedia Miner | API that provides a simplified access to Wikipedia and models its structure semantically [Milne et al. 2008] *http://sourceforge.net/ projects/wikipedia-miner/* |
| WikiPrep | A Perl tool for preprocessing Wikipedia XML dumps [Gabrilovich and Markovitch 2007] *http://www.cs.technion.ac.il/ ~gabr/resources/ code/wikiprep/* |
| W.H.A.T. Wikipedia | An analytic tool for Wikipedia with two main functionalities: an article network and extensive statistics. It contains a visualization of the article |
| Hybrid Analysis Tool | networks and a powerful interface to analyze the behavior of authors. *http://sourceforge.net/ projects/ w-h-a-t/* |

| **Wikipedia mining demos** | |
|---|---|
| DBpedia Online Access | Online access of DBpedia data (103M facts extracted from Wikipedia) via a SPARQL query endpoint and as Linked Data. [Auer et al. 2007] *http://wiki.dbpedia.org/ OnlineAccess* |
| YAGO | Demo of the Yet Another Ontology YAGO, containing 1.7M entities and 14M facts [Suchanek et al. 2007] *http://www.mpii.mpg.de/~suchanek/yago* |
| QuALiM | A Question Answering system. Given a question in a natural language returns relevant passages from Wikipedia. [Kaisser 2008] *http://demos.inf.ed.ac.uk:8080/ qualim/* |
| Koru | A demo of a search interface that maps topics involved in both queries and documents to Wikipedia articles. Supports automatic and interactive query expansion. [Milne et al. 2007] *http://www.nzdl.org/koru* |
| Wikipedia Thesaurus | A large scale association thesaurus containing 78M associations [Nakayama et al. 2007a and 2008] *http://wikipedia-lab.org:8080/WikipediaThesaurusV2/* |
| Wikipedia English-Japanese dictionary | A dictionary returning translations from English into Japanese and vise versa, enriched with probabilities of these translations [Erdmann et al. 2007] *http://wikipedia-lab.org:8080/WikipediaBilingualDictionary/* |
| Wikify | Automatically annotates any text with links to Wikipedia articles [Mihalcea and Csomai 2007] *http://wikifyer.com/* |
| Wikifier | Automatically annotates any text with links to Wikipedia articles describing named entities *http://wikifier.labs.exalead.com/* |
| Location query server | Location data accessible via REST requests returning data in a SOAP envelope. Two requests are supported: A bounding box or a Wikipedia Article. The reply is the number of references made to locations within that bounding box, and a list of Wikipedia articles describing those locations. Or none, if the request is not a location. [Overell and Rüger 2006 and 2007] *http://www.doc.ic.ac.uk/~seo01/wiki/demos* |

| **Datasets** | |
|---|---|
| DBpedia | Facts extracted from Wikipedia infoboxes and link structure in RDF format. [Auer et al. 2007] *http://wiki.dbpedia.org* |
| Wikipedia Taxonomy | Taxonomy automatically generated from the network of categories in Wikipedia (RDF Schema format) [Ponzetto and Strube 2007; Zirn et al. 2008] *http://www.eml-research.de/ english/research/ nlp/download/ wikitaxonomy.php* |
| Semantic Wikipedia | A snapshot of Wikipedia automatically annotated with named entity tags. [Zaragossa et al. 2007] *http://www.yr-bcn.es/semanticWikipedia* |
| Cyc to Wikipedia | 50,000 automatically created mappings from Cyc terms to Wikipedia articles. [Medelyan and Legg 2008] |
| mappings | *http://www.cs.waikato.ac.nz/ ~olena/cyc.html* |
| Topic indexed documents | A set of 20 Computer Science technical reports indexed with Wikipedia articles as topics. 15 teams of 2 senior CS undergraduates have independently assigned topics from Wikipedia to each article. [Medelyan et al. 2008] *http://www.cs.waikato.ac.nz/ ~olena/wikipedia.html* |
| Locations in Wikipedia, ground truth | A manually annotated sample of 1000 Wikipedia articles. Each link in each article is annotated, whether it is a location or not. If yes, it contains the corresponding unique id from the TGN gazetteer. [Overell and Rüger 2006 and 2007] *http://www.doc.ic.ac.uk/ ~seo01/wiki/data_release* |

Table 8. Wikipedia tools and resources.

## 7. SUMMARY

A whole host of researchers have been quick to grasp the potential of Wikipedia as a resource for mining meaning: the literature is large and growing rapidly. We began this article by describing Wikipedia's creation process and structure (Section 2). The unique open editing philosophy, which accounts for its success, is subversive. Although regarded as suspect by the academic establishment, it is a remarkable concrete realization of the American pragmatist philosopher Peirce's proposal that knowledge be defined through its public character and future usefulness rather than any prior justification. Wikipedia is not just an encyclopedia but can be viewed as anything from a corpus, taxonomy, thesaurus, hierarchy of knowledge topics to a full-blown ontology. It includes explicit information about synonyms (redirects) and word senses (disambiguation pages), database-style information (infoboxes), semantic network information (hyperlinks), category information (category structure), discussion pages, and the full edit history of every article. Each of these sources of information can be mined in various ways.

Section 3 explains how Wikipedia is being exploited for natural language processing. Unlike WordNet, it was not created as a lexical resource that reflects the intricacies of human language. Instead, its primary goal is to provide encyclopedic knowledge across subjects and languages. However, the research described here demonstrates that it has, unexpectedly, immense potential as a repository of linguistic knowledge for natural language applications. In particular, its unique features allow well-defined tasks such as word sense disambiguation and word similarity to be addressed automatically—and the resulting level of performance is remarkably high. ESA [Gabrilovich and Markovitch, 2007] and Wikipedia Link-based Measure [Milne and Witten, 2008a], for example, take advantage of the extended and hyperlinked description of concepts that in WordNet were restricted to short glosses. Furthermore, whereas in WordNet the sense frequency was defined by a simple ranking of meaning, Wikipedia implicitly contains conditional probabilities of word meanings [Mihalcea and Csomai, 2007], which allows more accurate similarity computation and word sense disambiguation. While the current research in this area has been mostly restricted to English, the approaches are general enough to apply to other languages. Researchers on co-reference resolution and mining of multilingual information have only recently discovered Wikipedia; significant improvements in these areas can be expected. To our knowledge, its use as a resource for other tasks such as natural language generation, machine translation and discourse analysis, has not yet been explored. These areas are ripe for exploitation, and exciting discoveries can be expected.

Section 4 describes applications in information retrieval. New techniques for document classification and topic indexing make productive use of Wikipedia for searching and organizing document collections. These areas can take advantage of its unique properties while grounding themselves in—and building upon—existing research. In particular, document classification has gathered momentum and significant advances have been obtained over the state of the art. Question answering and entity ranking are less well addressed, because current techniques do not seem to take full advantage of Wikipedia—most simply treat it as just another corpus. We found little evidence of cross-pollination between this work and the information extraction efforts described in Section 5. Given how closely question answering and entity ranking depend on the extraction of facts and entities, we expect this to become a fruitful line of enquiry.

In Section 5 we turn to information extraction and ontology building; mining Wikipedia for topics, relations and facts and then organizing them into a single resource. This task is less well defined than those in Sections 3 and 4. Different researchers focus on different kinds of information: we have reviewed projects that identify movie directors and soccer players, composers, corporate descriptions and hierarchical and ontological relations. Techniques range from those developed for standard text corpora to ones that utilize Wikipedia-specific properties such as hyperlinks and the category structure. The extracted resources range in size from several hundred to several million relations, but the lack of a common basis for evaluation prevents any overall conclusions as to which approach performs best. We believe that an extrinsic evaluation would be most meaningful, and hope to see these systems compete on a well-defined task in an independent evaluation. It will also be interesting to see to what extent these resources are exploited by other research communities in the future.

Some authors have suggested using the Wikipedia editors themselves to perform ontology-building, an enterprise that might be thought of as mining Wikipedia's *people* rather than its *data*. Perhaps they understand the underlying driving force behind this massively successful resource better than most! Only time will tell whether the community is amenable to following such suggestions. The idea of moving to a more structured and ontologically principled Wikipedia raises an interesting question: how will it interact with the public, amateur-editor model? Does this signal the emergence of the Semantic Web? We suspect that, like the success of Wikipedia itself, the result will be something new, something that experts have not foreseen and may not condone. That is the glory of Wikipedia.

ACKNOWLEDGEMENTS

We warmly thank Evgeniy Gabrilovich, Rada Mihalcea, Dan Weld, Sören Auer, Fabian Suchanek and the YAGO team for their valuable comments on a draft of this paper. We are also grateful to Enrico Motta and Susan Wiedenbeck for guiding us in the right direction. Medelyan is supported by a scholarship from Google, Milne by the New Zealand Tertiary Education Commission.

.